\documentclass{article}
\usepackage[preprint]{neurips_2024}

\usepackage{amssymb}
\usepackage{latexsym}
\usepackage{amsmath}
\usepackage{dsfont}
\usepackage{fontawesome5}
\usepackage{wrapfig}

\usepackage{url}
\usepackage{xcolor}
\usepackage{graphicx}
\usepackage{booktabs}
\usepackage{array}
\usepackage{enumitem}
\usepackage{multirow}
\usepackage{multicol}
\usepackage{ulem}
\usepackage{caption}
\usepackage{hyperref}

\definecolor{newcolor}{rgb}{.8,.349,.1}

\author{
\textbf{Haoran Xi}$^{1,2}$ \quad
\textbf{Chen Liu}$^{3}$ \quad
\textbf{Xiaolin Li}$^{1,2}$ \\\\
$^1$Academy of Medical Engineering and Translational Medicine, Tianjin University, Tianjin, CHN\\ $^2$Hangzhou Institute of Medicine, Chinese Academy of Sciences, Hangzhou, CHN \\
$^3$ Yale University, New Haven, Connecticut, USA\\\\
Please direct correspondence to \url{xiaolinli@ieee.org}.\\
\faGithub{} The code is available at \href{https://github.com/ant1dote/FPGM.git}{\url{https://github.com/ant1dote/FPGM.git}}
}

\title{Frequency Prior Guided Matching: A Data Augmentation Approach for Generalizable Semi-Supervised Polyp Segmentation}

\makeatletter
\let\inserttitle\@title
\makeatother

\begin{document}
\maketitle

\begin{abstract}
Automated polyp segmentation is essential for early diagnosis of colorectal cancer, yet developing robust models remains challenging due to limited annotated data and significant performance degradation under domain shift. Although semi-supervised learning~(SSL) reduces annotation requirements, existing methods rely on generic augmentations that ignore polyp-specific structural properties, resulting in poor generalization to new imaging centers and devices. To address this, we introduce Frequency Prior Guided Matching~(FPGM), a novel augmentation framework built on a key discovery: polyp edges exhibit a remarkably consistent frequency signature across diverse datasets. FPGM leverages this intrinsic regularity in a two-stage process. It first learns a domain-invariant frequency prior from the edge regions of labeled polyps. Then, it performs principled spectral perturbations on unlabeled images, aligning their amplitude spectra with this learned prior while preserving phase information to maintain structural integrity. This targeted alignment normalizes domain-specific textural variations, thereby compelling the model to learn the underlying, generalizable anatomical structure. Validated on six public datasets, FPGM establishes a new state-of-the-art against ten competing methods. It demonstrates exceptional zero-shot generalization capabilities, achieving over 10\% absolute gain in Dice score in data-scarce scenarios. By significantly enhancing cross-domain robustness, FPGM presents a powerful solution for clinically deployable polyp segmentation under limited supervision.
\end{abstract}

% \linenumbers

\begin{figure}[!tb]
\centering
\begin{minipage}[c]{0.48\textwidth}
\includegraphics[width=\textwidth]{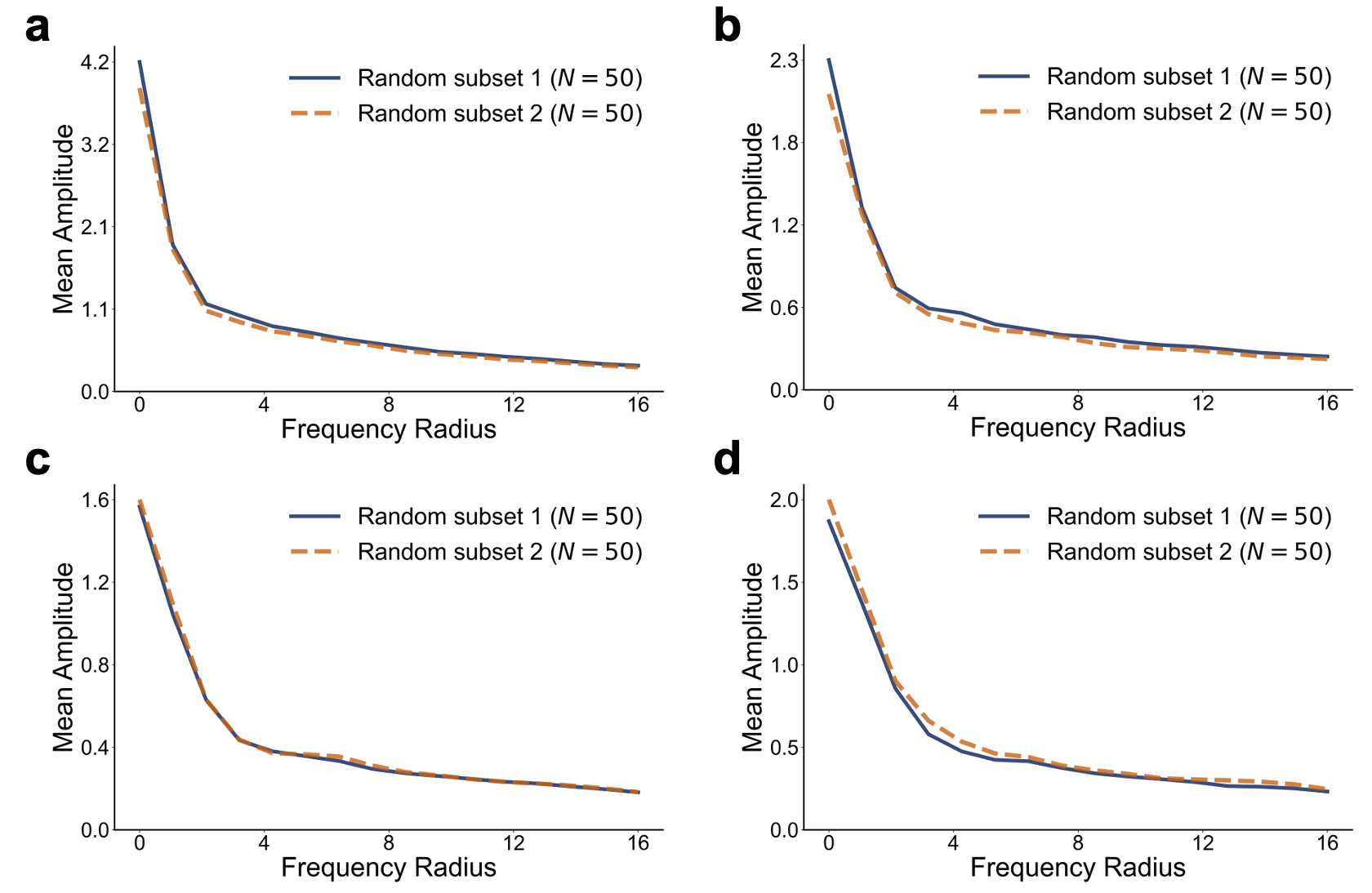}
\caption{\textbf{Average frequency signatures are remarkably similar between random subsets within the same dataset.} For four individual datasets, namely \textbf{(a)} \texttt{Kvasir}, \textbf{(b)} \texttt{CVC-ClinicDB}, \textbf{(c)} \texttt{CVC-ColonDB}, and \textbf{(d)} \texttt{ETIS}, two non-overlapping subsets of equal size are randomly sampled. The average radial frequency profile for each subset, derived from the Fast Fourier Transform~(FFT) of polyp edge regions, are highly consistent between the subsets.}
\label{fig:frequency_signature_separate}
\end{minipage}
\hfill
\begin{minipage}[c]{0.48\textwidth}
\includegraphics[width=\textwidth]{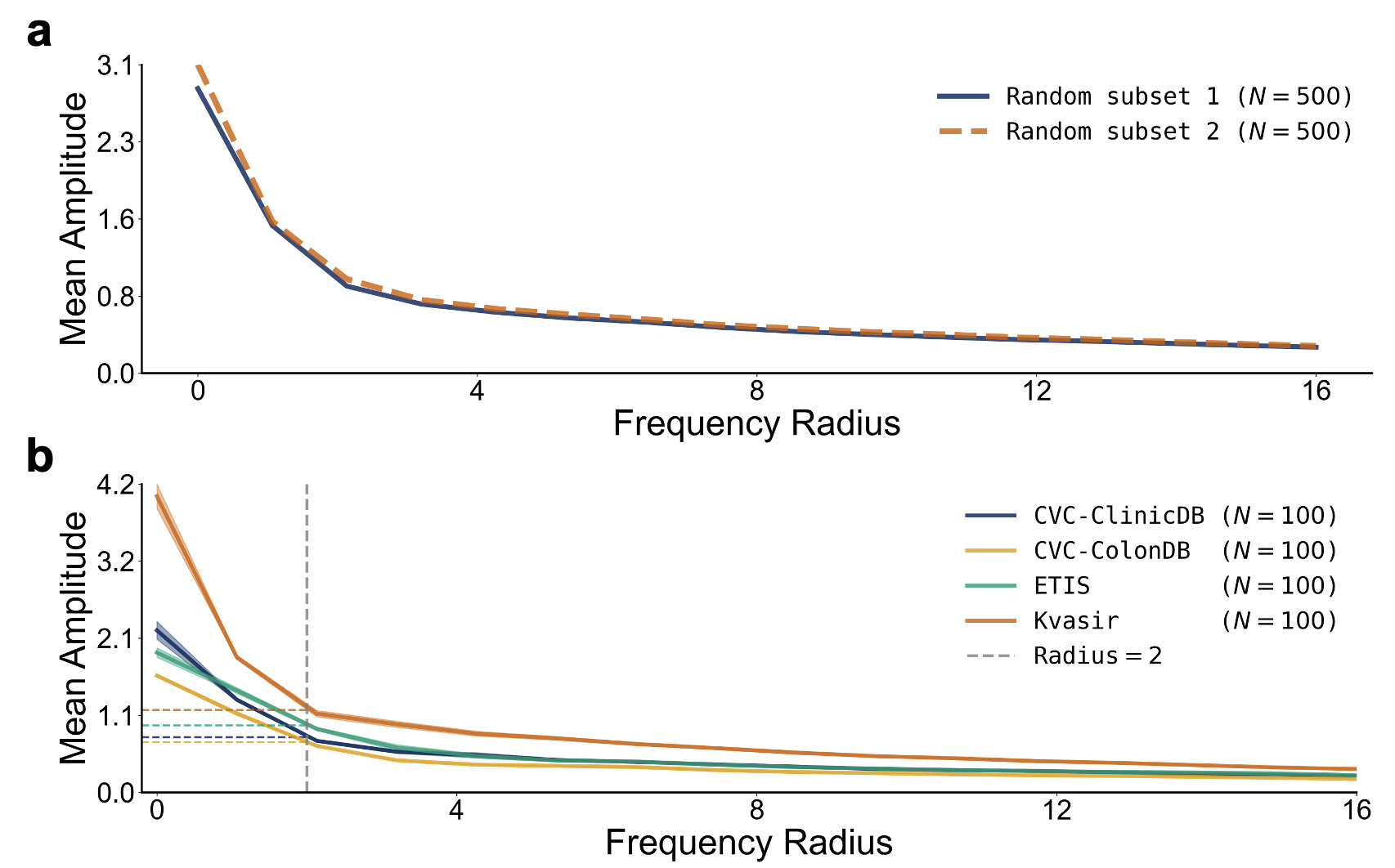}
\caption{\textbf{Average frequency signatures are similar across different datasets.} \textbf{(a)} In a mixture of the four datasets, two non-overlapping subsets of equal size are sampled and their profiles are shown. \textbf{(b)} The profile for each individual dataset is separately displayed, with a solid line indicating the mean and a shaded region indicating the standard deviation. Despite the domain diversity of the datasets, the frequency signatures are highly similar and consistent.}
\label{fig:frequency_signature_combined}
\end{minipage}
\end{figure}

\section{Introduction}
\label{sec:intro}

Colorectal cancer~(CRC) ranks among the most common and lethal malignancies worldwide~\citep{center2009worldwide}, and early detection of precancerous polyps is critical to reduce mortality~\citep{weitz2005colorectal}. Colonoscopy is the gold standard for polyp screening, but its diagnostic performance is highly operator-dependent, leading to a substantial rate of missed lesions~\citep{dekker2019colorectal}. Automated polyp segmentation using deep learning has emerged as a promising solution to support clinical decision-making~\citep{jain2023coinnet, cai2024know}. However, most existing methods rely on fully supervised learning and thus require large-scale, pixel-level annotations, which is a resource-intensive and time-consuming process in medical imaging~\citep{fan2020pranet, zhang2020adaptive}.

Semi-supervised learning (SSL) offers a compelling alternative by leveraging a small labeled dataset alongside a large pool of unlabeled images~\citep{wu2021collaborative}. A dominant strategy within SSL is consistency regularization, which encourages model predictions to remain stable under input perturbations~\citep{zhou2003learning}. This principle underlies many recent advances~\citep{sohn2020fixmatch, yang2023revisiting}, in which a weakly augmented image (e.g., flipping and rotation) is used to generate a pseudo-label that supervises a strongly augmented version (e.g., color jitter, Gaussian blur) of the same image. However, these generic perturbations are typically agnostic to the underlying anatomy. These random stylistic variations can obscure the intrinsic structural characteristics of polyps, causing models to overfit to superficial, domain-specific stylistic cues like illumination and color, rather than learning the generalizable structure itself. This issue is particularly evident under domain shift, where performance degrades significantly on data from different centers or endoscopic devices.

To address this, we move beyond such random perturbations and instead exploit a structural property inherent to colonoscopic polyp images. We empirically observe that radial frequency profiles, a proxy for textural characteristics, computed at polyp edges exhibit a consistent pattern. This pattern holds not only across randomly sampled subsets within a single dataset (Fig.~\ref{fig:frequency_signature_separate}) but also remains stable across datasets from different clinical centers (Fig.~\ref{fig:frequency_signature_combined}). This reproducible signature suggests the presence of a domain-invariant stylistic prior, which can serve as a powerful, biologically meaningful insight to guide the learning process and improve robustness.

Building on this insight, we propose Frequency Prior Guided Matching (FPGM), a data augmentation framework that learns this signature and uses it to perform guided style perturbations on unlabeled images. This approach allows us to normalize the stylistic representation of the unlabeled data against a learned canonical template, thereby reducing domain-specific interference and encouraging the model to focus on generalizable structure.
FPGM operates in two stages. First, it learns a one-dimensional frequency prior by aggregating the amplitude spectra of polyp edge regions across labeled images, as these spectra encode stylistic and textural information. Second, for each unlabeled image, it performs a principled perturbation by aligning its amplitude spectrum toward this prior, while preserving the phase to maintain structural integrity. This process, which we term guided style alignment, transfers domain-invariant knowledge from labeled to unlabeled samples in a structured manner.

\clearpage
\newpage
Our contributions are summarized as follows:
\begin{itemize}
\item We identified and validated a frequency prior that characterizes polyp edges in colonoscopic images. This prior is consistent across data sources and provides a quantifiable domain-invariant signal for structural regularization.
\item We proposed FPGM, a novel augmentation framework that leverages this observation via frequency prior matching. Our method performs targeted, structure-preserving perturbations on unlabeled images by aligning their frequency profiles to the learned template.
\item We demonstrated that FPGM achieved state-of-the-art performance through extensive experiments on six public datasets compared against ten recent competing methods. Remarkably, the performance of FPGM is substantially superior under domain shift, as verified on two external validation datasets. Our results established a new benchmark for generalizable semi-supervised polyp segmentation.
\end{itemize}

\section{Related Works}
\label{sec:related_work}

\subsection{Semi-supervised medical image segmentation}
Semi-supervised learning~(SSL) is crucial for leveraging vast quantities of unlabeled data, with consistency regularization being a central theme. A dominant approach is the teacher-student framework, where a student model learns from a more stable teacher. Numerous works have refined this dynamic. For instance, Dual-Student~\citep{ke2019dual} and ATSO~\citep{huo2021atso} were developed to address issues of tight model coupling and stagnant, ``lazy'' updates, respectively.

An alternative and powerful paradigm is co-training, which exploits diverse ``views'' from multiple models to provide mutual supervision. This principle is exemplified by Cross Pseudo Supervision~(CPS)~\citep{chen2021semi}, which uses two differently initialized networks to generate varied supervisory signals. This concept has also been adapted to operate within a single network, such as in MC-Net+~\citep{wu2022mutual} which enforces consistency across multiple parallel decoders, and in CUTS~\citep{liu2024cuts} which encourages embedding proximity among similar image patches.
A key refinement to consistency regularization involves modulating the learning process based on model uncertainty. For example, AC-MT~\citep{xu2023ambiguity} focuses on exploring information within highly uncertain regions. Similarly, URPC~\citep{luo2022semi} introduces a multi-scale uncertainty rectification to temper the consistency loss at outlier pixels, while CoraNet~\citep{shi2021inconsistency} also leverages pixel-wise inconsistency as a proxy for uncertainty to guide its self-training strategy.

Beyond these core paradigms, several recent efforts have explored alternative strategies to improve semi-supervised segmentation. Some focus on enhancing pseudo-label reliability. For instance, ComWin~\citep{wu2023compete} adopts a ``compete-to-win'' scheme to select the most confident predictions. Others incorporate contrastive learning, such as the task-agnostic framework by Basak et al.~\cite{basak2023pseudo}, or RCPS~\citep{zhao2023rcps} which integrates contrastive objectives with rectified pseudo-supervision. New forms of regularization have also emerged, including adversarial consistency to bridge labeled and unlabeled domains~\citep{lei2022semi}, and task-level constraints that complement data-level augmentations, as in DTC~\citep{luo2021semi}. DiffKillR~\citep{liu2025diffkillr} introduces diffeomorphic transformations to guide segmentation under limited supervision. Additionally, Huang et al.~\cite{huang2025rethinking} proposes a copy-paste strategy to mitigate confirmation bias introduced by CutMix~\citep{yun2019cutmix}. In contrast to these methods, which emphasize model uncertainty or diverse views, our approach introduces a novel consistency regularization mechanism driven by a learned frequency prior.

\subsection{Polyp segmentation}

Numerous deep learning methods have been specifically tailored for polyp segmentation, each addressing distinct challenges of the task. Some works focus on enhancing feature representation and data diversity. For instance, Guo et al.~\cite{guo2020learn} introduced a confidence-guided manifold mixup to combat data scarcity and class imbalance. MSRF-Net~\citep{srivastava2021msrf} improves feature propagation by exchanging information across multi-scale receptive fields to generate more accurate segmentation maps.

A significant trend involves improving the delineation of ambiguous polyp boundaries. FuzzyNet~\citep{patel2022fuzzynet} employs a fuzzy attention module to concentrate on challenging pixels near the boundary. This focus is shared by CFA-Net~\citep{zhou2023cross}, which incorporates features from an auxiliary boundary prediction network, and MEGANet~\citep{bui2024meganet}, which fuses a classical Laplacian operator with attention to suppress boundary noise. Similarly, UACANet~\citep{kim2021uacanet} explicitly models an uncertainty map to refine its predictions. Another notable approach, SANet~\citep{wei2021shallow}, contributes by not only using a shallow attention module to filter background noise but also by introducing a color exchange operation to decouple content from style. 

Some transformer-based polyp segmentation methods also exhibit competitive performance by modeling the long-range dependencies, such as Polyp-PVT~\citep{dong2021polyp}, CTNet~\citep{xiao2024ctnet}, and FCN-transformer~\citep{sanderson2022fcn}.

\subsection{Image perturbation method}
Data perturbation is a cornerstone of modern deep learning, designed to enrich training data and enhance model generalization. A significant body of work focuses on perturbations in the spatial domain. Foundational techniques include Mixup~\citep{zhang2017mixup}, which linearly interpolates between image pairs, and methods like Cutout~\citep{devries2017improved} or CutMix~\citep{yun2019cutmix}, which improve robustness by swapping rectangular image patches. More advanced strategies build on this, such as BCPNet~\citep{bai2023bidirectional}, which employs bidirectional copy-pasting to align distributions, or AugMix~\citep{hendrycks2019augmix}, which composes multiple simple augmentations into a stochastically generated sample.

A distinct and emerging line of work explores perturbations in the frequency domain~\citep{wang2023dfm}, operating on the principle that an image's style is primarily encoded in its amplitude spectrum, while its structure lies in the phase. FACT~\citep{xu2021fourier}, for instance, exchanges amplitude spectra between images to synthesize style-diversified samples and reduce the domain gap. Similarly, FSDR~\citep{huang2021fsdr} leverages the Discrete Cosine Transform to decompose images into domain-invariant and domain-variant components, then randomizes the latter to disrupt superficial style cues.

While these methods validate the utility of frequency-space perturbations, they typically rely on inter-image style swapping or the randomization of broadly defined domain-variant components. In contrast, our FPGM introduces a novel, intra-domain regularization strategy guided by a learned, task-specific prior. Instead of exchanging spectra between arbitrary image pairs, FPGM first establishes a domain-invariant frequency template by aggregating knowledge from the labeled data. It then perturbs unlabeled samples by gently aligning their spectral profiles towards this stable, guiding template. This represents a more controlled and targeted approach to bias neutralization, as the perturbation is not random, but rather a deliberate and gentle shift towards a statistically robust, domain-average representation.

\section{Frequency Prior Guided Matching}
\label{sec:method}
\begin{figure}[!tb]
\centering
\includegraphics[width=\textwidth]{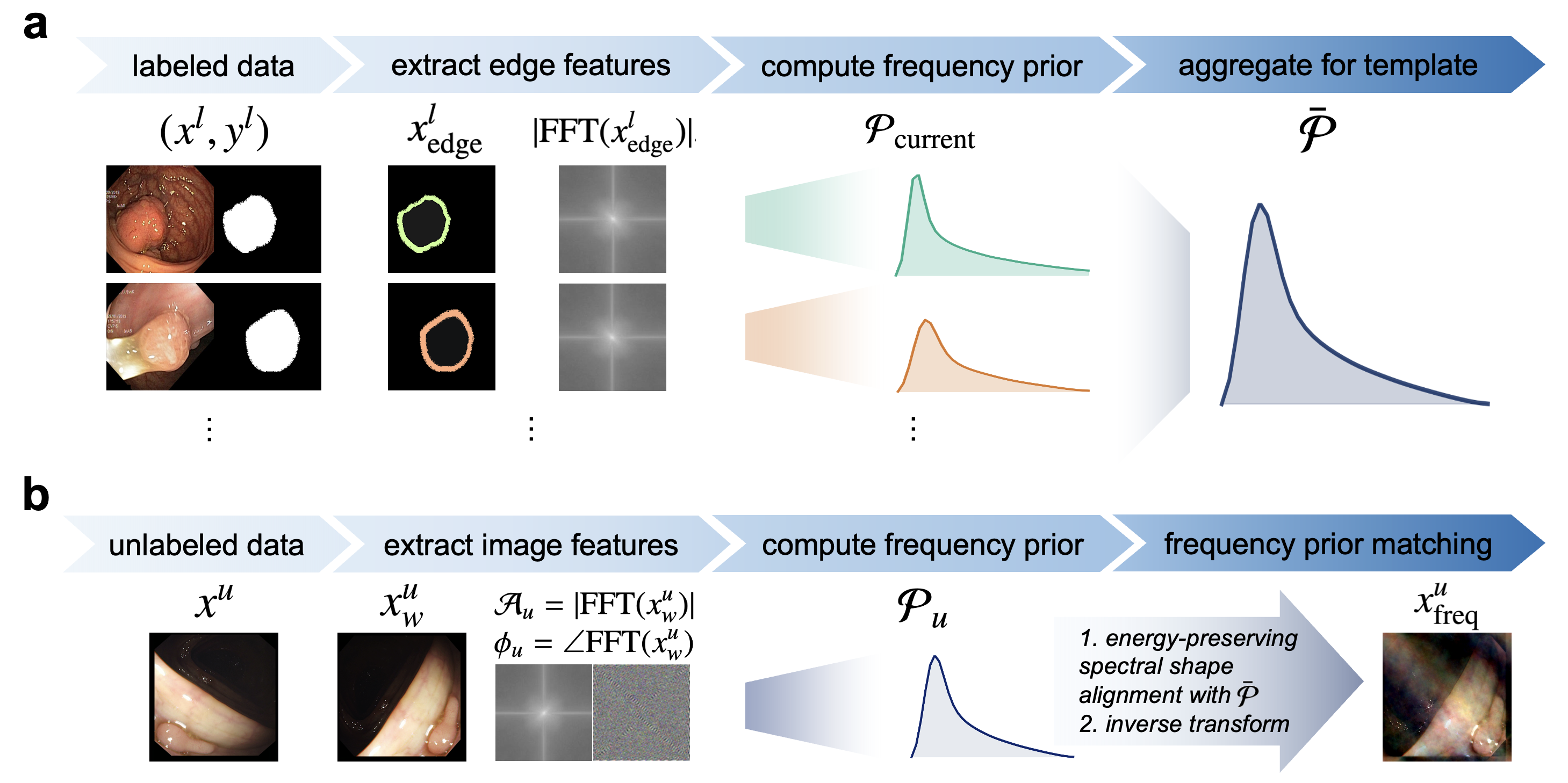}
\caption{\textbf{Schematic of Frequency Prior Guided Matching~(FPGM)}. \textbf{(a)} In Stage I~(Section~\ref{sec:method_stage1}), we learn the frequency prior from labeled data. We extract the polyp boundary and compute its one-dimensional radial frequency profile, which characterizes the sample's frequency signature. These signatures are then aggregated across labeled samples to obtain a stable and smooth global frequency prior. \textbf{(b)} In Stage II~(Section~\ref{sec:method_stage2}), we perform spectral shape alignment to augment unlabeled data. Each unlabeled image is transformed into the frequency domain and decomposed into amplitude and phase. Its radial amplitude profile is aligned towards the frequency prior from Stage I. The resulting profile is reconstructed into the two-dimensional Fourier spectrum under an energy preservation constraint. After combining with the original phase, it is transformed back to the image domain to generate the augmented sample.}
\label{fig:schematic}
\end{figure}

We propose \textbf{Frequency Prior Guided Matching~(FPGM)}, a two-stage framework that enforces consistency during training and guides the model to learn invariant, generalizable features~(Fig.~\ref{fig:schematic}). FPGM operates in two key stages:
\begin{enumerate}
\item \textbf{Stage I}~(Fig.~\ref{fig:schematic}a): Learning a stable, one-dimensional frequency prior $\bar{\mathcal{P}}$ from the labeled data, capturing the spectral signature of polyp edge regions.
\item \textbf{Stage II}~(Fig.~\ref{fig:schematic}b): Using this learned prior to perturb unlabeled images through spectral shape alignment, thereby generating structured, domain-invariant augmentations.
\end{enumerate}

\subsection{Problem setup and overview}
\label{sec:method_setup}
We consider a semi-supervised segmentation setting. Let $x \in \mathbb{R}^{H \times W \times 3}$ denote an input colonoscopy image and $y \in \{0,1\}^{H \times W}$ its corresponding binary mask, with 1 being the foreground and 0 being the background. The training dataset consists of a small labeled subset $\mathcal{D}_L = \{(x_i^l, y_i^l)\}_{i=1}^{N_L}$ and a large unlabeled subset $\mathcal{D}_U = \{x_j^u\}_{j=1}^{N_U}$. The full dataset is therefore $\mathcal{D} = \mathcal{D}_L \cup \mathcal{D}_U$, where the number of labeled samples is significantly limited, i.e., $N_L \ll N_U$.

Our final goal is to train a segmentation model $f_\theta$ with parameters $\theta$ that predicts segmentation masks from input images and performs robustly across domains, which we achieve through our proposed FPGM.

\subsection{\textbf{Stage I:} Learning the frequency prior}
\label{sec:method_stage1}

Given a labeled data pair $(x^l, y^l)$ consisting of an image and the corresponding ground-truth segmentation mask, we derive a global frequency profile $\bar{\mathcal{P}}$ that characterizes the radial frequency spectrum of polyp boundary regions. This frequency prior captures the spectral characteristics inherent to polyp edges, and will serve as a template for alignment in the data augmentation process.

\paragraph{Edge detection}

For each ground-truth mask $y^l$, we first compute the gradient map $\mathcal{G}$ which highlights the polyp boundary, as formalized in Eqn~\eqref{eqn:edge_detection}.
\begin{equation}
\label{eqn:edge_detection}
\mathcal{G} = \sqrt{(y^l * S_x)^2 + (y^l * S_y)^2}
\end{equation}
Here, $S_x$ and $S_y$ are respectively the horizontal and vertical Sobel kernels~\citep{kanopoulos1988design}, and $*$ denotes 2D convolution. To improve robustness, we further apply morphological dilation to the resulting gradient map, resulting in a binary edge mask $\mathcal{M}_\text{edge} \in \{0,1\}^{H \times W}$ that precisely delineates the polyp boundary.

\paragraph{Spectral analysis of edge regions}

We extract the textural content at polyp boundaries by first converting the input image $x^l$ to grayscale $x_\text{gray}^l$ and then isolating edge-specific information, as described in Eqn~\eqref{eqn:edge_region}, where $\odot$ denotes the Hadamard product.
\begin{equation}
\label{eqn:edge_region}
x_\text{edge}^l = x_\text{gray}^l \odot \mathcal{M}_\text{edge}
\end{equation}
The frequency domain representation is obtained via the Fast Fourier Transform~(FFT), as shown in Eqn~\eqref{eqn:edge_fft}, where $|\cdot|$ computes the amplitude spectrum.
\begin{equation}
\label{eqn:edge_fft}
\widehat{x}_\text{edge} = |\text{FFT}(x_\text{edge}^l)|
\end{equation}

\paragraph{Radial profile computation}

To ensure rotation invariance, we extract a radial profile by averaging spectral magnitudes at each radial distance $r$ from the center of the frequency domain. This process is formulated in Eqn~\eqref{eqn:radial_profile}.
\begin{equation}
\label{eqn:radial_profile}
\mathcal{P}_\text{current}(r) = \frac{1}{|\Omega_r|} \sum_{(u,v) \in \Omega_r} \widehat{x}_\text{edge}(u,v)
\end{equation}
Here, $\Omega_r = \bigl\{ (u,v) : \sqrt{u^2 + v^2} = r \Bigr\}$ represents the set of frequency coordinates at the radial distance $r$.

This operation reduces the two-dimensional~(2D) spectral information to a one-dimensional~(1D) rotation-invariant energy distribution curve $\mathcal{P}_\text{current}$.

\paragraph{Frequency prior template from aggregation}

Since the frequency profiles of individual samples might exhibit noise, we aggregate the individual priors into a stable global frequency prior $\bar{\mathcal{P}}$ representative of the entire dataset. The aggregation process employs an exponential moving average~(EMA), as described in Eqn~\eqref{eqn:aggregation}.
\begin{equation}
\label{eqn:aggregation}
\bar{\mathcal{P}} \leftarrow (1 - \mu) \cdot \mathcal{P}_\text{current} + \mu \cdot \bar{\mathcal{P}}
\end{equation}
Here, $\mu \in [0,1]$ is the momentum hyperparameter that controls the update rate, which defaults to $\mu = 0.999$. This accumulation strategy enables $\bar{\mathcal{P}}$ to progressively integrate spectral statistics from all labeled samples, resulting in a robust frequency-domain characterization of polyp boundaries.

The complete process of Stage I is illustrated in Fig.~\ref{fig:schematic}a.

\subsection{\textbf{Stage II:} Spectral shape alignment for domain-invariant augmentation}
\label{sec:method_stage2}

The learned frequency prior $\bar{\mathcal{P}}$ encapsulates a consistent, domain-invariant spectral signature of polyp edges, serving as a canonical template for knowledge transfer from labeled to unlabeled data. The core innovation of our FPGM framework lies in steering the spectral characteristics of unlabeled images toward this template through a principled process termed spectral shape alignment. Unlike naive spectrum mixing approaches, our method meticulously decouples spectral properties into shape and energy components, performs alignment in normalized shape space, and reconstructs the signal to ensure targeted, structure-preserving perturbations.

Starting from a weakly-augmented unlabeled image $x^u_w$, the spectral shape alignment process unfolds through four steps.

\paragraph{Frequency decomposition}

We transform $x^u_w$ into the frequency domain using the FFT, decomposing it into amplitude and phase spectra, as described in Eqn~\eqref{eqn:frequency_decomposition}.
\begin{equation}
\label{eqn:frequency_decomposition}
\mathcal{A}_u = |\text{FFT}(x^u_w)|, \quad \phi_u = \angle\text{FFT}(x^u_w)
\end{equation}
Here, $|\cdot|$ and $\angle(\cdot)$ respectively denote magnitude and phase extraction from the complex-valued FFT output. The amplitude spectrum $\mathcal{A}_u$ primarily encodes stylistic and textural information, while the phase spectrum $\phi_u$ preserves critical spatial structure. Our strategy selectively modifies only the amplitude while preserving the phase to maintain structural integrity.

\paragraph{Spectral shape and energy characterization}

The 2D amplitude spectrum $\mathcal{A}_u$ is condensed into a more tractable 1D representation by quantifying its radial profile $\mathcal{P}_u$ using the radial profile computation technique from Section~\ref{sec:method_stage1}. A critical distinction must be made regarding the computational scope: while the prior $\bar{\mathcal{P}}$ is selectively derived from the polyp edge regions of the labeled data to capture a ``pure'' polyp signal, here $\mathcal{P}_u$ encompasses the amplitude spectrum of the entire unlabeled image.

Consequently, $\mathcal{P}_u$ represents the holistic spectral signature of the complete image $x^u_w$, encapsulating the mixed frequency content from potential targets~(polyps) and surrounding context~(normal tissue and artifacts). To enable principled alignment of this multiplexed profile, we characterize $\mathcal{P}_u$ in two fundamental aspects:
\begin{enumerate}[itemsep=2pt,topsep=4pt]
    \item[(i)] Total energy $E_u$~(Eqn~\eqref{eqn:total_energy}): the overall intensity of frequency components, quantified using the L1-norm.
    \begin{equation}
    \label{eqn:total_energy}
        E_u = \sum_r \mathcal{P}_u(r)
    \end{equation}
    \item[(ii)] Normalized shape $||\mathcal{P}_u||$~(Eqn~\eqref{eqn:normalized_shape}): a representation of energy distribution across frequencies, independent of total energy.
    \begin{equation}
    \label{eqn:normalized_shape}
        ||\mathcal{P}_u|| = \frac{\mathcal{P}_u}{E_u + \epsilon}
    \end{equation}
    Here, $\epsilon$ is a small constant for numerical stability by preventing division by zero. $||\mathcal{P}_u||$ can be interpreted as a probability distribution over different frequency radii.
\end{enumerate}

\paragraph{Shape-space alignment}

This step constitutes the heart of our knowledge transfer mechanism from labeled to unlabeled data in FPGM. We perform guided interpolation in normalized shape space rather than on raw spectral profiles. The learned template $\bar{\mathcal{P}}$ is first normalized to obtain its shape representation $||\bar{\mathcal{P}}||$, and a perturbed shape $||\mathcal{P}_\text{pert}||$ is synthesized through guided interpolation as described in Eqn~\eqref{eqn:guided_interpolation}, where $\gamma \in [0,1]$ controls the guidance strength, which we empirically set to $\gamma = 0.05$.
\begin{equation}
\label{eqn:guided_interpolation}
||\mathcal{P}_{\text{pert}}|| = (1 - \gamma) \cdot ||\mathcal{P}_{u}|| + \gamma \cdot ||\bar{\mathcal{P}}||
\end{equation}
This operation effectively pulls the spectral shape of the unlabeled image toward the canonical, domain-invariant polyp edge prior $\bar{\mathcal{P}}$.

\paragraph{Energy-preserving reconstruction}

To finalize the perturbed spectral profile, we rescale the generated shape $||\mathcal{P}_\text{pert}||$ by the original total energy $E_u$ to ensure energy preservation, as shown in Eqn~\eqref{eqn:energy_preservation}.
\begin{equation}
\label{eqn:energy_preservation}
\mathcal{P}_\text{pert} = ||\mathcal{P}_\text{pert}|| \cdot E_u
\end{equation}
This energy-preserving step ensures that perturbations redistribute spectral energy according to the new shape without altering the overall energy content. The 1D profile $\mathcal{P}_\text{pert}$ is subsequently broadcast to a 2D amplitude spectrum $\mathcal{A}_\text{pert}$ based on the frequency radius, as shown in Eqn~\eqref{eqn:broadcast}.
\begin{equation}
\label{eqn:broadcast}
\mathcal{A}_\text{pert}(u, v) = \mathcal{P}_\text{pert} \Bigl( r(u, v) \Bigr)
\end{equation}
Finally, the perturbed image $x^u_\text{freq}$ is reconstructed by combining the new amplitude spectrum with the original phase spectrum, followed by the inverse FFT. The reconstruction process is described in Eqn~\eqref{eqn:reconstruction}.
\begin{equation}
\label{eqn:reconstruction}
x_{\text{freq}}^u = \text{FFT}^{-1} \Bigl( \mathcal{A}_{\text{pert}} \cdot e^{i \phi_u} \Bigr)
\end{equation}

This spectral shape alignment pipeline modifies the image's spectral ``style'' to align with the learned prior while preserving fundamental structure, thereby facilitating robust and generalizable feature learning.

The complete process of Stage II is illustrated in Fig.~\ref{fig:schematic}b.

\subsection{Training framework and loss function}
\label{sec:method_training}

Our framework integrates supervised and unsupervised learning objectives through a unified training paradigm. The loss function $\mathcal{L}$ combines a supervised loss $\mathcal{L}_\text{sup}$ with two complementary unsupervised consistency losses, $\mathcal{L}_\text{unsup}$ and $\mathcal{L}_\text{freq}$, as shown in Eqn~\eqref{eqn:loss_function}.
\begin{equation}
\label{eqn:loss_function}
\mathcal{L} = \mathcal{L}_\text{sup} + \lambda_\text{unsup} \cdot \mathcal{L}_\text{unsup} + \lambda_\text{freq} \cdot \mathcal{L}_\text{freq}
\end{equation}
Here, $\lambda_\text{unsup}$ and $\lambda_\text{freq}$ are scalar weights that balance the contribution of each unsupervised loss term. Empirically, we set both weights to 0.5, ensuring equal emphasis on conventional augmentation and our FPGM counterpart.

\paragraph{Supervised learning component} 

For each labeled data pair $(x^l, y^l)$, the model $f_\theta$ produces a prediction $f_\theta(x^l)$. The supervised loss employs a balanced combination of Cross-Entropy and Dice loss to address both pixel-wise classification accuracy and region-based segmentation quality~\citep{taghanaki2019combo}, as shown in Eqn~\eqref{eqn:loss_supervised}.
\begin{equation}
\label{eqn:loss_supervised}
\mathcal{L}_\text{sup} = \frac{1}{2} \biggl(\text{CrossEntropy} \Bigl(f_\theta(x^l), y^l \Bigr) + \text{Dice} \Bigl( f_\theta(x^l), y^l \Bigr) \biggr)
\end{equation}

\paragraph{Unsupervised learning component}
For unlabeled images, we enforce consistency between predictions on weakly-augmented and strongly-augmented views through a dual-consistency framework, following prior works in this field~\citep{sohn2020fixmatch, yang2023revisiting}.

Given an unlabeled image $x^u$, we first apply weak augmentation~(e.g., random flip) to obtain $x^u_w$. High-confidence pseudo-labels $y^u$ are generated by filtering the model's predictions based on confidence thresholding, as described in Eqn~\eqref{eqn:pseudo_label}.
\begin{equation}
\label{eqn:pseudo_label}
y^u = \text{argmax} \biggl(f_\theta(x^u_w) \odot \mathds{I} \Bigl( \text{Conf}\bigl(f_\theta(x^u_w) \bigr) \geq \tau_\text{c} \Bigr) \biggr)
\end{equation}
Here, $\text{Conf}\bigl(f_\theta(x^u_w) \bigr)$ represents pixel-wise prediction confidence, $\mathds{I}$ is the indicator function that filters out low-confidence pixels, $\odot$ is the Hadamard product, and $\tau_c = 0.95$ is the confidence threshold. For generality of notation, the final step is expressed as an argmax operation, which assigns the predicted class for each pixel. In case of binary classification, it can be replaced by a simple binarization.

Next, we generate two distinct strongly-augmented views from the same weakly-augmented image $x^u_w$. The first view, $x^u_s$, is produced using a standard strong augmentation policy~(e.g., color jitter, Gaussian blur). The second view, $x^u_\text{freq} = \text{FPGM}(x^u_w)$, is generated using our proposed spectral shape alignment method. Because these augmentation strategies are predefined, any spatial transformation they introduce~(denoted $\mathcal{T}_w^u$ and $\mathcal{T}_\text{freq}^u$) can be applied to the pseudo label $y^u$ to maintain image-label consistency. In practice, FPGM operates on individual pixel intensities without altering spatial structure, so its corresponding transformation is simply the identity: $\mathcal{T}_\text{freq}^u = \mathbb{I}$.

The model's predictions on these views, $f_\theta(x_s^u)$ and $f_\theta(x_\text{freq}^u)$, are both optimized using the pseudo-label $y^u$, yielding two consistency losses shown in Eqn~\eqref{eqn:loss_unsupervised} and \eqref{eqn:loss_FPGM}.

\begin{minipage}{0.47\textwidth}
\begin{equation}
\label{eqn:loss_unsupervised}
\mathcal{L}_\text{unsup} = \text{Dice} \Bigl(f_\theta(x^u_s), \mathcal{T}_w^u(y^u) \Bigr)
\end{equation}
\end{minipage}
\hfill
\begin{minipage}{0.47\textwidth}
\begin{equation}
\label{eqn:loss_FPGM}
\mathcal{L}_\text{freq} = \text{Dice} \Bigl(f_\theta(x^u_\text{freq}), \mathcal{T}_\text{freq}^u(y^u) \Bigr)
\end{equation}
\end{minipage}

This dual-consistency framework encourages the model to maintain invariance to both generic visual variations and targeted spectral perturbations, promoting robust feature learning across different domains.

\section{Main Empirical Results}
\label{sec:experiments}

\begin{wraptable}{r}{0.55\textwidth}
\centering
\vskip -12pt
\caption{Summary and data partitioning of the six public datasets used in our evaluation. The dash symbol indicates the corresponding datasets are used as held-out unseen data to evaluate generalization.}
\label{tab:datasets}
\resizebox{0.48\textwidth}{!}{%
\begin{tabular}{lcccc}
\toprule
Dataset & Sample size & Train & Validation & Test \\
\midrule
\texttt{Kvasir}       & 1000 & 700 & 100 & 200  \\
\texttt{CVC-ClinicDB} & 612  & 400 & 62  & 150  \\
\texttt{CVC-ColonDB}  & 380  & 230 & 50  & 100  \\
\texttt{ETIS}         & 196  & 100 & 32  & 64   \\
\texttt{CVC-300}      & 60   & --- & --- & 60   \\
\texttt{BKAI}         & 1000 & --- & --- & 1000 \\
\bottomrule
\end{tabular}
}
\vskip -8pt
\end{wraptable}

This section presents a comprehensive experimental evaluation of our proposed FPGM framework. We first describe the datasets and experimental settings, followed by quantitative and qualitative comparisons with cutting-edge semi-supervised segmentation methods under varying supervision levels. Finally, we assess the generalization capability on completely unseen datasets to validate domain adaptation performance.

\subsection{Experimental setup}

\paragraph{Datasets}

We performed experiments on six public colonoscopy datasets: \texttt{Kvasir}~\citep{jha2019Kvasir}, \texttt{CVC-ClinicDB}~\citep{bernal2015wm}, \texttt{CVC-ColonDB}~\citep{jain2023coinnet}, \texttt{ETIS}~\citep{silva2014toward}, \texttt{CVC-300}~\citep{bernal2012towards}, and \texttt{BKAI}~\citep{bkai-igh-neopolyp}. Detailed dataset statistics are provided in Table~\ref{tab:datasets}.

Our experimental design evaluates two critical aspects: 
\begin{enumerate}
    \item[(i)] Standard in-domain semi-supervised segmentation performance on four datasets~(\texttt{Kvasir}, \texttt{CVC-ClinicDB}, \texttt{CVC-ColonDB}, and \texttt{ETIS}), and
    \item[(ii)] Zero-shot domain generalization capability on two completely unseen datasets~(\texttt{CVC-300} and \texttt{BKAI}) that remain excluded from all training and validation phases.
\end{enumerate} 
\vskip -16pt

\paragraph{Data partitioning and sampling protocol}

Following established protocols~\citep{fan2020pranet, wang2025dynamic}, we constructed training and validation sets by sampling from the first four datasets listed in Table~\ref{tab:datasets}. Specifically, we randomly pooled 1,430 samples for training and 244 samples for validation from \texttt{Kvasir}, \texttt{CVC-ClinicDB}, \texttt{CVC-ColonDB}, and \texttt{ETIS}. The model performance is evaluated separately on the unseen test set of these four datasets.

To simulate varying supervision levels, we created three distinct splits from the unified training pool of 1,430 samples: 5\%~(70 samples), 10\%~(140 samples), and 20\%~(280 samples) serve as labeled sets~($\mathcal{D}_L$), with the remaining samples forming the unlabeled sets~($\mathcal{D}_U$) in each scenario.
\vskip -16pt

\paragraph{Baseline methods}

To demonstrate the effectiveness and robustness of our proposed FPGM method, we compare it with recent SOTA methods, including URPC~(MedIA'22)~\citep{luo2022semi}, BSNet~(TMI'23)~\citep{he2023bilateral}, AC-MT~(MedIA'23)~\citep{xu2023ambiguity}, CauSSL~(ICCV'23)~\citep{miao2023caussl}, BCPNet~(CVPR'23)~\citep{bai2023bidirectional}, UniMatch~(CVPR'23)~\citep{yang2023revisiting}, ABD~(CVPR'24)~\citep{chi2024adaptive}, DyCON~(CVPR'25)~\citep{assefa2025dycon}, CVBM~(TIP'25)~\citep{cao2025background} and PICK~(IJCV'25)~\citep{zeng2025pick}.

\paragraph{Image preprocessing}

All images and masks were resized to $256 \times 256$ resolution with pixel values normalized to [0, 1]. Data were stored in H5 format for efficient processing. Experiments were conducted on a workstation equipped with Intel Core i9-10900K CPU and NVIDIA RTX 3090 GPU running Ubuntu 20.04, implemented in Python 3.8.8 and PyTorch 1.10.0 with CUDA 11.5.
\vskip -16pt

\paragraph{Implementation details}

For a fair comparison, all models in our experiments employ the same UNet backbone~\citep{2015U}. Optimization was performed by the Stochastic Gradient Descent~(SGD) optimizer with a momentum of 0.9 and a weight decay of 0.0001. The initial learning rate was set to 0.01 and followed a polynomial decay schedule. Each training batch comprised 8 labeled and 8 unlabeled images. Each model was trained for a total of 30,000 iterations. For reproducibility, all experiments were performed under the same random seeds. All competing methods were re-implemented and evaluated under the exact same experimental settings.
\vskip -16pt

\paragraph{Evaluation metrics}
To evaluate segmentation performance, we
adopted four standard segmentation metrics: Dice-S{\o}rensen coefficient~(Dice), Jaccard index~(Jaccard), Hausdorff distance 95th percentile~(HD95) and average symmetric surface distance~(ASD).
\vskip -2pt

\begin{table}[!tb]
\centering
\caption{Semi-supervised segmentation performances on four public datasets under various supervision levels. The best results are \textbf{bolded}.}
\label{tab:comparison}
\resizebox{\textwidth}{!}{%
\begin{tabular}{lll cccc cccc cccc cccc}
\toprule

\multirow{2}{*}{\textbf{Setting}} & \multirow{2}{*}{\textbf{Method}} & \multirow{2}{*}{\textbf{Venue}} & \multicolumn{4}{c}{\texttt{Kvasir}} & \multicolumn{4}{c}{\texttt{CVC-ClinicDB}} & \multicolumn{4}{c}{\texttt{CVC-ColonDB}} & \multicolumn{4}{c}{\texttt{ETIS}} \\
\cmidrule(lr){4-7} \cmidrule(lr){8-11} \cmidrule(lr){12-15} \cmidrule(lr){16-19}
& & & Dice$\uparrow$ & Jaccard$\uparrow$ & HD95$\downarrow$ & ASD$\downarrow$ & Dice$\uparrow$ & Jaccard$\uparrow$ & HD95$\downarrow$ & ASD$\downarrow$ & Dice$\uparrow$ & Jaccard$\uparrow$ & HD95$\downarrow$ & ASD$\downarrow$ & Dice$\uparrow$ & Jaccard$\uparrow$ & HD95$\downarrow$ & ASD$\downarrow$ \\
\midrule
\multirow{11}{*}{5\% labeled} &
URPC     & MedIA'22 & 48.56 & 36.64 & 76.79 & 30.41 & 34.48 & 25.89 & 57.87 & 22.77 & 12.80 & 8.77  & 53.62 & 26.23 & 22.68 & 17.86 & 53.79 & 30.40 \\
& BSNet    & TMI'23   & 41.10 & 29.99 & 86.14 & 37.86 & 19.53 & 14.21 & 64.81 & 28.07 & 7.60  & 5.18  & 48.21 & 25.82 & 20.18 & 15.73 & 69.92 & 42.53 \\
& AC-MT    & MedIA'23 & 42.20 & 30.21 & 85.63 & 37.82 & 36.25 & 27.22 & 73.71 & 32.79 & 20.03 & 13.50 & 76.74 & 40.71 & 25.40 & 19.50 & 89.40 & 49.03 \\
& CauSSL   & ICCV'23  & 43.89 & 32.23 & 80.94 & 35.11 & 31.01 & 23.35 & 62.04 & 27.96 & 10.73 & 6.78  & 57.51 & 26.70 & 24.11 & 18.00 & 74.65 & 44.24 \\
& BCPNet   & CVPR'23  & 42.07 & 30.80 & 82.07 & 34.76 & 26.51 & 20.02 & 61.96 & 26.06 & 10.57 & 7.00  & 60.26 & 27.73 & 6.44  & 4.20  & 64.21 & 31.11 \\
& UniMatch & CVPR'23  & 74.57 & 64.48 & 42.79 & 14.26 & 67.24 & 59.24 & 41.31 & 12.58 & 45.34 & 38.13 & 55.57 & 23.87 & 45.44 & 37.77 & 46.28 & 23.86 \\
& ABD      & CVPR'24  & 44.14 & 30.97 & 79.60 & 38.04 & 36.31 & 25.52 & 69.54 & 32.65 & 21.83 & 14.38 & 79.79 & 43.44 & 23.48 & 15.71 & 75.49 & 40.33 \\
& DyCON    & CVPR'25  & 65.85 & 55.12 & 54.14 & 18.60 & 49.75 & 42.53 & 49.59 & 16.77 & 40.77 & 33.24 & 46.84 & 18.02 & 36.04 & 28.87 & 48.25 & 23.76 \\
& CVBM     & TIP'25   & 48.35 & 37.74 & 69.55 & 27.34 & 32.10 & 25.07 & 57.11 & 22.68 & 21.41 & 15.97 & 53.05 & 22.10 & 12.17 & 8.65  & 87.20 & 54.31 \\
& PICK     & IJCV'25  & 59.42 & 46.63 & 73.40 & 30.72 & 47.25 & 36.26 & 65.87 & 27.74 & 35.40 & 25.92 & 90.18 & 44.45 & 26.22 & 19.17 & 81.96 & 40.12 \\
\cmidrule(lr){2-19}
& \textbf{FPGM~(ours)} & --- & \textbf{80.66} & \textbf{71.34} & \textbf{36.80} & \textbf{12.27} & \textbf{78.01} & \textbf{70.47} & \textbf{27.41} & \textbf{7.46} &  \textbf{63.13} & \textbf{54.63} & \textbf{40.79} & \textbf{13.47} & \textbf{51.07} & \textbf{43.69} & \textbf{39.64} & \textbf{19.02} \\

\midrule
\multirow{11}{*}{10\% labeled} &
URPC     & MedIA'22 & 46.31 & 36.92 & 53.37 & 18.20 & 39.64 & 31.02 & 37.54 & 9.22 & 15.37 & 10.80 & 36.43 & 13.66 & 10.39 & 7.22 & 41.91 & 18.69 \\
& BSNet    & TMI'23   & 37.44 & 29.99 & 73.61 & 28.55 & 30.04 & 22.71 & 52.26 & 20.29 & 7.47  & 4.83  & 74.00 & 37.70 & 8.40  & 6.18 & 96.35 & \textbf{6.71} \\
& AC-MT    & MedIA'23 & 41.79 & 31.84 & 66.05 & 24.45 & 39.99 & 30.38 & 50.66 & 17.05 & 19.62 & 14.24 & 59.73 & 26.33 & 13.62 & 9.49 & 68.54 & 35.10 \\
& CauSSL   & ICCV'23  & 42.58 & 32.43 & 60.50 & 21.65 & 32.18 & 23.50 & 50.60 & 16.74 & 19.73 & 14.64 & 47.58 & 20.73 & 13.03 & 9.50 & 58.46 & 26.78 \\
& BCPNet   & CVPR'23  & 46.32 & 35.22 & 61.30 & 21.70 & 37.67 & 30.07 & 67.58 & 29.05 & 24.73 & 19.20 & 54.54 & 20.73 & 13.18 & 9.84 & 83.02 & 47.64 \\
& UniMatch & CVPR'23  & 77.13 & 67.86 & 40.03 & 11.31 & 76.75 & 69.20 & 26.78 & \textbf{5.98} & 63.03 & 54.66 & 35.96 & 11.88 & 38.98 & 33.88 & \textbf{33.13} & 13.82 \\
& ABD      & CVPR'24  & 41.79 & 29.07 & 75.46 & 37.71 & 36.75 & 26.03 & 74.91 & 36.19 & 24.12 & 15.83 & 81.98 & 45.66 & 22.51 & 15.84 & 95.60 & 53.54 \\
& DyCON    & CVPR'25  & 68.55 & 57.77 & 50.31 & 15.74 & 67.93 & 59.46 & 39.74 & 10.48 & 47.05 & 39.16 & 44.77 & 13.89 & 24.42 & 19.72 & 63.42 & 32.81 \\
& CVBM     & TIP'25   & 62.60 & 51.37 & 54.97 & 18.06 & 48.26 & 39.91 & 47.67 & 12.88 & 29.39 & 22.83 & 45.94 & 16.25 & 17.00 & 12.61 & 51.95 & 25.59 \\
& PICK     & IJCV'25  & 66.36 & 56.09 & 46.43 & 14.17 & 66.44 & 57.28 & 42.30 & 11.84 & 48.96 & 39.56 & 57.72 & 23.77 & 36.58 & 29.02 & 72.66 & 32.92 \\
\cmidrule(lr){2-19}
& \textbf{FPGM~(ours)} & --- & \textbf{82.24} & \textbf{73.21} & \textbf{31.74} & \textbf{9.44} & \textbf{79.83} & \textbf{72.36} & \textbf{24.59} & 6.00 & \textbf{68.03} & \textbf{59.55} & \textbf{30.16} & \textbf{8.76} & \textbf{44.84} & \textbf{38.46} & 39.52 & 18.08 \\

\midrule
\multirow{11}{*}{20\% labeled} &
URPC     & MedIA'22 & 65.06 & 55.05 & 46.24 & 14.85 & 55.59 & 47.03 & 40.32 & 9.73 & 38.62 & 31.29 & 43.50 & 15.60 & 32.35 & 26.74 & 34.01 & 14.54 \\
& BSNet    & TMI'23   & 58.39 & 48.61 & 50.00 & 15.62 & 47.31 & 39.41 & 53.78 & 17.50 & 37.08 & 30.03 & 45.45 & 19.59 & 21.13 & 17.21 & 40.77 & 19.23 \\
& AC-MT    & MedIA'23 & 59.32 & 48.56 & 54.68 & 18.47 & 49.48 & 40.54 & 44.08 & 12.77 & 33.93 & 26.60 & 54.62 & 21.26 & 29.70 & 24.00 & 37.17 & 17.29 \\
& CauSSL   & ICCV'23  & 60.32 & 49.35 & 50.72 & 16.29 & 52.82 & 44.29 & 50.13 & 15.79 & 33.14 & 25.79 & 53.26 & 25.59 & 23.93 & 19.31 & 42.55 & 24.16 \\
& BCPNet   & CVPR'23  & 61.19 & 50.38 & 52.91 & 16.93 & 52.38 & 43.38 & 51.03 & 17.21 & 35.75 & 28.18 & 65.01 & 27.80 & 23.74 & 19.04 & \textbf{33.04} & \textbf{14.35} \\
& UniMatch & CVPR'23  & 80.88 & 71.35 & 36.05 & 11.31 & 75.42 & 67.92 & 27.83 & 6.85 & 68.93 & 59.34 & 41.85 & 13.66 & 53.00 & 45.31 & 40.28 & 17.96 \\
& ABD      & CVPR'24  & 45.09 & 32.51 & 62.38 & 27.94 & 36.66 & 26.84 & 62.40 & 25.56 & 25.65 & 18.18 & 70.00 & 33.79 & 24.94 & 17.98 & 78.51 & 43.14 \\
& DyCON    & CVPR'25  & 79.11 & 69.30 & 42.19 & 13.69 & 74.20 & 65.38 & 36.51 & 9.35 & 59.34 & 50.22 & 54.76 & 19.82 & 51.27 & 42.52 & 55.76 & 26.75 \\
& CVBM     & TIP'25   & 72.22 & 61.82 & 43.10 & 13.46 & 64.75 & 56.16 & 39.37 & 10.72 & 44.84 & 37.45 & 49.08 & 18.49 & 38.58 & 31.45 & 50.21 & 20.47 \\
& PICK     & IJCV'25  & 74.94 & 64.39 & 46.27 & 15.69 & 67.62 & 59.21 & 31.71 & 8.24 & 53.77 & 45.10 & 55.63 & 19.09 & 53.69 & 43.78 & 53.04 & 22.71 \\
\cmidrule(lr){2-19}
& \textbf{FPGM~(ours)} & ---& \textbf{84.65} & \textbf{76.65} & \textbf{29.79} & \textbf{8.06} & \textbf{82.48} & \textbf{75.19} & \textbf{25.45} & \textbf{5.65} & \textbf{72.19} & \textbf{63.15} & \textbf{35.30} & \textbf{11.89} & \textbf{58.83} & \textbf{51.08} & 42.49 & 17.26 \\
\bottomrule
\end{tabular}
}
\end{table}

\subsection{In-domain semi-supervised segmentation performance}

Table~\ref{tab:comparison} presents the comparative performance of our proposed FPGM against ten representative contemporary models. The results establish FPGM as the new state-of-the-art (SOTA) method for semi-supervised polyp segmentation. Its dominance is particularly evident in the region-based metrics (Dice and Jaccard), where it consistently ranks first across all datasets and supervision levels. This highlights the method's unparalleled effectiveness in ensuring complete and accurate polyp delineation. In addition, FPGM demonstrates highly competitive results in the boundary-sensitive metrics (HD95 and ASD). For example, on the challenging \texttt{CVC-ColonDB} dataset, our method achieves substantial reductions in boundary errors, validating that our frequency-guided perturbations allow the model to capture more precise structural details.

\begin{figure}[!tb]
\centering
\includegraphics[trim={0.2cm 0 0.8cm 0}, clip, width=0.98\textwidth]{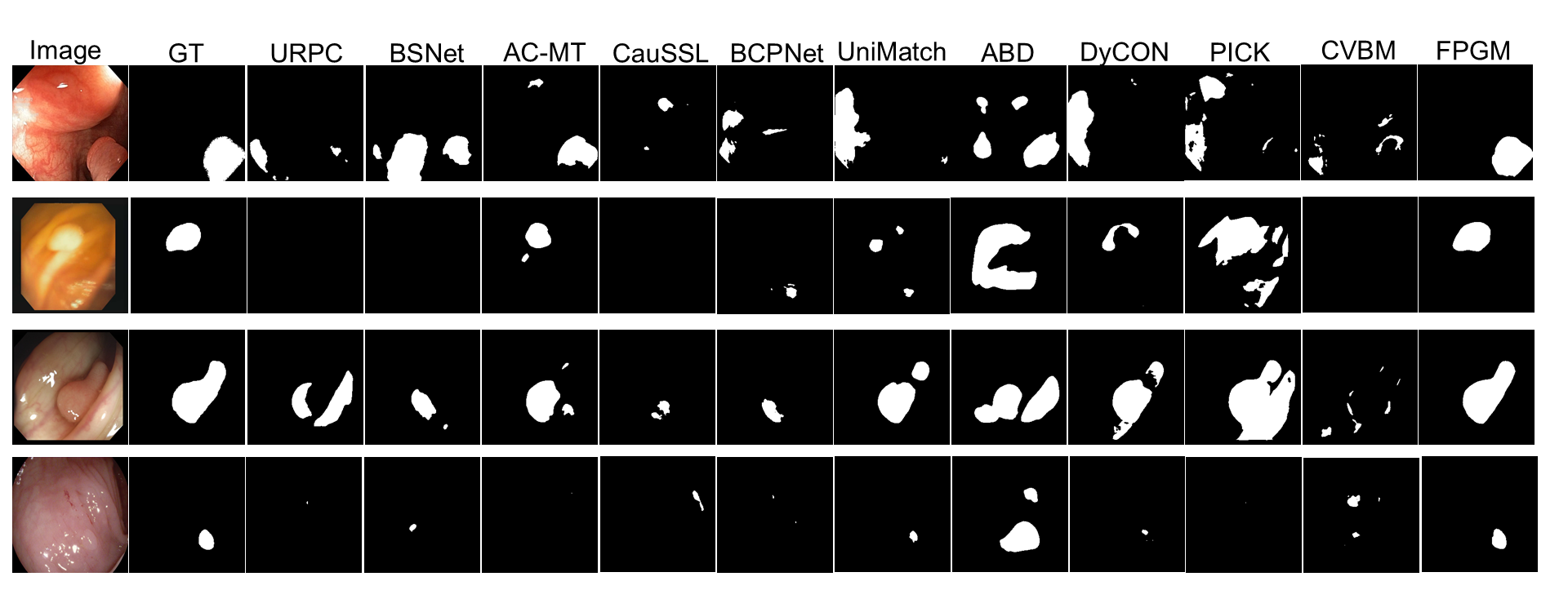}
\caption{Qualitative comparison of semi-supervised segmentation results on four public datasets using \textbf{5\% labeled data}. From top to bottom, each row displays a representative example from the \texttt{Kvasir}, \texttt{CVC-ClinicDB}, \texttt{CVC-ColonDB}, and \texttt{ETIS} datasets, respectively.}
\label{fig:comparison_5pct}
\end{figure}

\begin{figure}[!tb]
\centering
\includegraphics[width=0.98\textwidth]{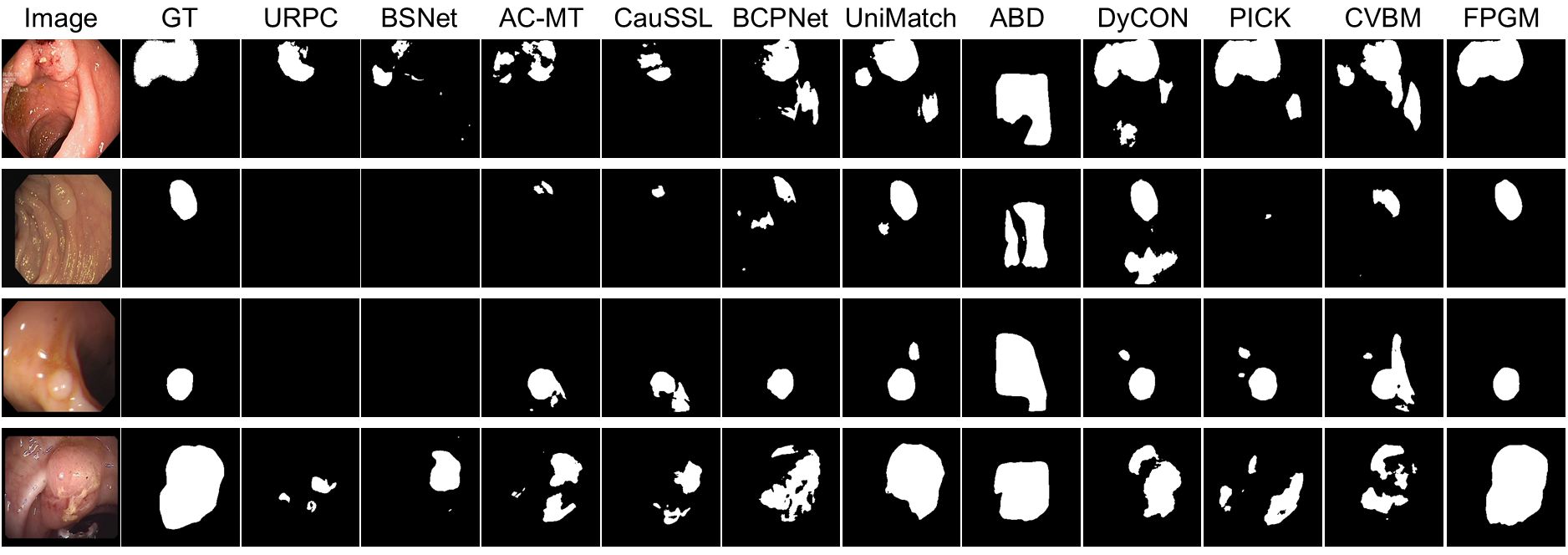}
\caption{Qualitative comparison of semi-supervised segmentation results on four public datasets using \textbf{10\% labeled data}. From top to bottom, each row displays a representative example from the \texttt{Kvasir}, \texttt{CVC-ClinicDB}, \texttt{CVC-ColonDB}, and \texttt{ETIS} datasets, respectively.}
\label{fig:comparison_10pct}
\end{figure}

Qualitative results under 5\% supervision~(Fig.~\ref{fig:comparison_5pct}) demonstrate that FPGM generates significantly more accurate and robust segmentation masks with high fidelity to ground truth. In contrast, baseline methods frequently exhibit false negatives~(missing polyps, rows 2 and 4), false positives~(noisy artifacts, row 1), or incomplete segmentation~(partial polyp capture, row 3). The results under 10\% supervision~(Fig.~\ref{fig:comparison_10pct}) further confirm the effectiveness of our approach across challenging cases, including large diffusely-bordered polyps and small subtle lesions. FPGM consistently produces segmentation masks with superior completeness and boundary accuracy compared to competitors.

The results show that the performance advantage of FPGM extends to all supervision levels, from low-data regimes to higher supervision settings. Although most of the baseline methods improve with increased labeled data, the performance gap between FPGM and the second-best method remains substantial in all settings. This demonstrates that our frequency-guided perturbation strategy provides effective regularization against domain-specific overfitting, regardless of the amount of labeled data available.

\begin{table}[!t]
\centering
\caption{Zero-shot generalization performance on \textbf{completely unseen datasets}. Models were trained on the previous four public datasets~(\texttt{Kvasir}, \texttt{CVC-ClinicDB}, \texttt{CVC-ColonDB} and \texttt{ETIS}) using 5\%, 10\%, or 20\% labeled data, and \textbf{directly evaluated without re-training} on \texttt{CVC-300} and \texttt{BKAI}. The best results are \textbf{bolded}.}
\label{tab:comparison_generalization}
\resizebox{\textwidth}{!}{%
\begin{tabular}{l l cccc cccc cccc cccc cccc cccc}
\toprule

\multirow{2}{*}{\textbf{Method}} & \multirow{2}{*}{\textbf{Venue}} & \multicolumn{8}{c}{\textbf{5\% labeled data}} & \multicolumn{8}{c}{\textbf{10\% labeled data}} & \multicolumn{8}{c}{\textbf{20\% labeled data}} \\
\cmidrule(lr){3-10} \cmidrule(lr){11-18} \cmidrule(lr){19-26}

& & \multicolumn{4}{c}{\texttt{CVC-300}~(Unseen)} & \multicolumn{4}{c}{\texttt{BKAI}~(Unseen)} & \multicolumn{4}{c}{\texttt{CVC-300}~(Unseen)} & \multicolumn{4}{c}{\texttt{BKAI}~(Unseen)} & \multicolumn{4}{c}{\texttt{CVC-300}~(Unseen)} & \multicolumn{4}{c}{\texttt{BKAI}~(Unseen)} \\
\cmidrule(lr){3-6} \cmidrule(lr){7-10} \cmidrule(lr){11-14} \cmidrule(lr){15-18} \cmidrule(lr){19-22} \cmidrule(lr){23-26}

& & Dice$\uparrow$ & Jaccard$\uparrow$ & HD95$\downarrow$ & ASD$\downarrow$ & Dice$\uparrow$ & Jaccard$\uparrow$ & HD95$\downarrow$ & ASD$\downarrow$ & Dice$\uparrow$ & Jaccard$\uparrow$ & HD95$\downarrow$ & ASD$\downarrow$ & Dice$\uparrow$ & Jaccard$\uparrow$ & HD95$\downarrow$ & ASD$\downarrow$ & Dice$\uparrow$ & Jaccard$\uparrow$ & HD95$\downarrow$ & ASD$\downarrow$ & Dice$\uparrow$ & Jaccard$\uparrow$ & HD95$\downarrow$ & ASD$\downarrow$ \\
\midrule
URPC     & MedIA'22 & 3.33  & 1.95  & 18.73 & 9.85  & 29.97 & 21.71 & 72.38 & 35.02 & 11.26 & 8.30  & 14.45 & 5.50  & 25.29 & 19.97 & 38.54 & 17.23 & 51.95 & 44.07 & 20.89 & 8.17  & 47.15 & 39.05 & 39.30 & 16.14 \\
BSNet    & TMI'23   & 13.10 & 7.90  & 19.82 & 12.27 & 21.62 & 15.10 & 75.40 & 38.81 & 12.49 & 8.80  & 40.17 & 22.47 & 15.65 & 11.41 & 86.39 & 49.44 & 27.74 & 23.11 & 22.59 & 12.45 & 36.28 & 29.66 & 43.52 & 17.99 \\
AC-MT    & MedIA'23 & 29.33 & 20.86 & 71.31 & 34.69 & 24.97 & 17.40 & 83.21 & 42.39 & 21.63 & 16.42 & 46.06 & 25.96 & 26.66 & 20.11 & 61.93 & 30.63 & 30.04 & 24.18 & 24.62 & 11.48 & 37.80 & 30.36 & 41.82 & 17.12 \\
CauSSL   & ICCV'23  & 22.04 & 12.20 & 22.56 & 12.56 & 27.14 & 18.88 & 76.14 & 37.58 & 24.47 & 19.40 & 38.66 & 20.91 & 24.87 & 18.91 & 50.19 & 22.96 & 36.45 & 29.89 & 37.78 & 19.48 & 41.32 & 33.10 & 42.88 & 17.52 \\
BCPNet   & CVPR'23  & 8.34  & 4.75  & 27.34 & 17.83 & 20.83 & 14.86 & 73.46 & 35.55 & 34.95 & 27.75 & 41.02 & 21.00 & 29.01 & 21.80 & 63.05 & 31.20 & 46.31 & 39.03 & 48.83 & 22.56 & 45.29 & 36.97 & 59.53 & 29.24 \\
UniMatch & CVPR'23  & 64.17 & 57.18 & 33.54 & 13.40 & 55.62 & 46.33 & 46.95 & 20.63 & 84.07 & 76.42 & 16.50 & 6.29  & 62.73 & 54.17 & 35.23 & 14.35 & 83.52 & 76.31 & 20.85 & 6.64  & 60.61 & 52.10 & 35.65 & 14.44 \\
ABD      & CVPR'24  & 15.67 & 10.53 & 41.91 & 23.00 & 30.45 & 20.68 & 77.25 & 39.06 & 25.01 & 15.84 & 79.69 & 44.25 & 29.62 & 19.74 & 77.68 & 40.08 & 39.30 & 27.81 & 65.23 & 29.81 & 31.93 & 22.74 & 59.70 & 29.69 \\
DyCON    & CVPR'25  & 60.81 & 53.79 & 26.66 & 9.06  & 48.64 & 40.05 & 48.73 & 20.19 & 67.32 & 59.67 & 18.31 & 5.88  & 52.79 & 43.22 & 57.41 & 24.67 & 72.95 & 66.21 & 12.60 & 4.07  & 61.78 & 52.68 & 42.03 & 17.39 \\
CVBM     & TIP'25   & 27.37 & 20.44 & 32.78 & 15.75 & 27.87 & 21.22 & 94.52 & 51.61 & 37.41 & 30.80 & 22.22 & 8.24  & 42.37 & 34.01 & 50.50 & 21.73 & 67.70 & 58.53 & 36.62 & 14.32 & 51.91 & 43.58 & 41.26 & 17.51 \\
PICK     & IJCV'25  & 61.43 & 50.97 & 58.79 & 24.98 & 39.36 & 29.21 & 80.05 & 38.63 & 72.05 & 62.44 & 53.10 & 20.79 & 51.11 & 42.79 & 44.02 & 18.19 & 73.84 & 64.84 & 37.90 & 13.52 & 53.06 & 44.18 & 45.23 & 18.81 \\
\midrule
\textbf{FPGM~(ours)} & ---& \textbf{82.46} & \textbf{75.42} & \textbf{21.00} & \textbf{7.09} & \textbf{65.84} & \textbf{56.64} & \textbf{34.69} & \textbf{13.37} & \textbf{89.55} & \textbf{81.87} & \textbf{12.30} & \textbf{4.54} & \textbf{68.68} & \textbf{59.91} & \textbf{28.11} & \textbf{10.56} & \textbf{86.80} & \textbf{80.89} & \textbf{8.14} & \textbf{2.85} & \textbf{69.53} & \textbf{60.94} & \textbf{28.23} & \textbf{9.90} \\
\bottomrule
\end{tabular}
}
\end{table}

\begin{figure}[!t]
\centering
\includegraphics[width=\textwidth]{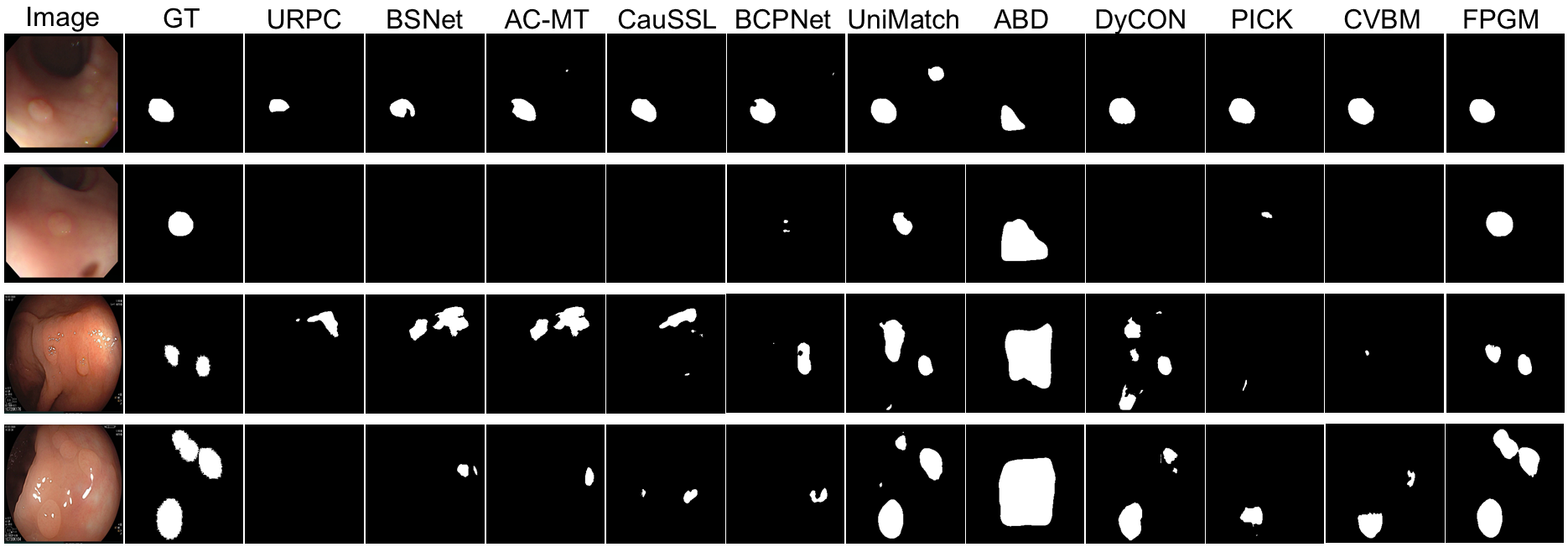}
\caption{Qualitative comparison of the zero-shot generalization capability on \textbf{completely unseen datasets} for models trained using 10\% labeled data and \textbf{directly evaluated without re-training}. The models were tested on \texttt{CVC-300}~(top two rows) and \texttt{BKAI}~(bottom two rows).}
\label{fig:comparison_generalization}
\end{figure}

\subsection{Zero-shot domain generalization performance}

To rigorously assess the zero-shot domain generalization capability, we directly evaluated models without re-training on two completely unseen datasets: \texttt{CVC-300} and \texttt{BKAI}. These models were trained using data from the four aforementioned datasets under 5\%, 10\%, and 20\% supervision. Table~\ref{tab:comparison_generalization} summarizes the performance of FPGM against all baselines on these unseen datasets.

The results consistently demonstrate FPGM's exceptional generalization ability. Across all supervision levels, our method achieves substantially higher scores on both unseen datasets compared to all competing methods. The performance gap is particularly pronounced; for instance, at 5\% supervision, the Dice score of FPGM surpasses the second-best method~(UniMatch) by 18.29\% (82.46\% vs. 64.17\%) on \texttt{CVC-300} and by 10.22\% (65.84\% vs. 55.62\%) on \texttt{BKAI}. The robust performance on unseen domains strongly suggests that by learning invariance to domain-specific frequency characteristics, our model successfully captures intrinsic, generalizable structural properties of polyps rather than superficial domain-specific cues.

An interesting observation in Table~\ref{tab:comparison_generalization} shows that FPGM achieved slightly higher Dice scores on \texttt{CVC-300} with 10\% supervision~(89.55\%) compared to 20\% supervision~(86.80\%). We attribute this minor fluctuation to statistical variance associated with the small \texttt{CVC-300} test set and the stochastic nature of data sampling when creating labeled training pools. The specific subset in the 10\% split may have been particularly beneficial in generalizing to the \texttt{CVC-300} domain. Importantly, across all settings, our method consistently and substantially outperforms competing approaches under fair comparisons, validating the overall robustness and superiority of our framework.

Qualitative results on the unseen datasets~(Fig.~\ref{fig:comparison_generalization}) also provide compelling evidence of generalization capability. For example, FPGM is the only method capable of identifying small, subtle lesions from \texttt{CVC-300}~(second row), while all competing methods fail. When confronted with multiple polyps in \texttt{BKAI} dataset~(third and fourth rows), FPGM accurately segments each instance, whereas baselines produce fragmented masks or merge distinct lesions. These challenging cases demonstrate that through learning invariance to stable frequency priors, our model successfully captures intrinsic and generalizable polyp characteristics, leading to robust performance on out-of-distribution data.

\section{In-depth Analyses}

This section provides comprehensive analyses of our proposed FPGM framework through ablations studies on loss components, hyperparameter sensitivity, and specificity of the learned frequency prior. We examine the individual contributions of each component and demonstrate the principled design choices underlying our approach.

\subsection{Ablation studies on FPGM loss components}
To systematically assess the effectiveness of our proposed framework, we conducted comprehensive ablation studies under 5\% and 10\% labeled data settings, with results detailed in Table~\ref{tab:ablation_seen}. The analysis reveals several critical insights that demonstrate the superiority and complementary nature of our frequency-guided consistency loss $\mathcal{L}_\text{freq}$.

\subsubsection{In-domain loss component ablation study}

As demonstrated in Table~\ref{tab:ablation_seen}, our frequency-guided regularization alone~($\mathcal{L}_\text{sup}+\mathcal{L}_\text{freq}$) systematically outperforms the standard semi-supervised baseline~($\mathcal{L}_\text{sup}+\mathcal{L}_\text{unsup}$) across all datasets and supervision levels. This consistent advantage is particularly pronounced in the 5\% low-data regime and robustly maintained in the 10\% setting. The results strongly suggest that our regularization strategy, which transfers knowledge via a stable structural prior, is inherently more effective than conventional consistency methods based on standard augmentations.

\begin{table}[!t]
\centering
\caption{Ablation study on the loss function components using the four public datasets. The best results are \textbf{bolded}.}
\label{tab:ablation_seen}
\resizebox{\textwidth}{!}{%
\begin{tabular}{llcccccccccccccccc}
\toprule
\multirow{2}{*}{\textbf{Setting}} & \multirow{2}{*}{\textbf{Design}} & \multicolumn{4}{c}{\texttt{Kvasir}} & \multicolumn{4}{c}{\texttt{CVC-ClinicDB}} & \multicolumn{4}{c}{\texttt{CVC-ColonDB}} & \multicolumn{4}{c}{\texttt{ETIS}} \\
\cmidrule(lr){3-6} \cmidrule(lr){7-10} \cmidrule(lr){11-14} \cmidrule(lr){15-18}
& & Dice$\uparrow$ & Jaccard$\uparrow$ & HD95$\downarrow$ & ASD$\downarrow$ & Dice$\uparrow$ & Jaccard$\uparrow$ & HD95$\downarrow$ & ASD$\downarrow$ & Dice$\uparrow$ & Jaccard$\uparrow$ & HD95$\downarrow$ & ASD$\downarrow$ & Dice$\uparrow$ & Jaccard$\uparrow$ & HD95$\downarrow$ & ASD$\downarrow$ \\
\midrule
\multirow{4}{*}{5\% labeled} & $\mathcal{L}_\text{sup}$ & 60.41 & 47.37 & 87.07 & 35.24 & 51.33 & 41.24 & 74.88 & 31.05 & 33.95 & 24.93 & 109.30 & 51.65 & 35.11 & 26.80 & 103.20 & 49.69 \\
& $\mathcal{L}_\text{sup} + \mathcal{L}_\text{unsup}$ & 74.35 & 64.13 & 41.86 & 14.30 & 69.60 & 61.12 & 36.94 & 10.45 & 48.38 & 41.12 & \textbf{31.94} & \textbf{9.80} & 39.67 & 33.15 & 46.16 & 24.49 \\
& $\mathcal{L}_\text{sup} + \mathcal{L}_\text{freq}$ & 77.14 & 66.94 & 46.44 & 15.06 & 72.72 & 63.95 & 41.07 & 14.64 & 56.13 & 47.03 & 62.41 & 25.04 & 48.86 & 41.54 & 41.42 & 19.38 \\
& $\mathcal{L}_\text{sup}+\mathcal{L}_\text{unsup}+\mathcal{L}_\text{freq}$ & \textbf{80.66} & \textbf{71.34} & \textbf{36.80} & \textbf{12.27} & \textbf{78.01} & \textbf{70.47} & \textbf{27.41} & \textbf{7.46} & \textbf{63.13} & \textbf{54.63} & 40.79 & 13.47 & \textbf{51.07} & \textbf{43.69} & \textbf{39.64} & \textbf{19.02} \\
\midrule
\multirow{4}{*}{10\% labeled} & $\mathcal{L}_\text{sup}$ & 72.65 & 61.26 & 52.13 & 17.71 & 73.92 & 64.08 & 49.62 & 16.95 & 49.14 & 39.65 & 59.40 & 24.40 & 33.62 & 27.27 & 67.52 & 38.97 \\
& $\mathcal{L}_\text{sup} + \mathcal{L}_\text{unsup}$ & 78.51 & 69.03 & 35.18 & 10.91 & 76.59 & 69.22 & \textbf{24.55} & \textbf{5.45} & 63.28 & 55.00 & 34.23 & 9.62 & 35.85 & 31.78 & \textbf{33.67} & 17.26 \\
& $\mathcal{L}_\text{sup} + \mathcal{L}_\text{freq}$ & 80.80 & 71.91 & 32.65 & \textbf{9.28} & 78.13 & 70.71 & 30.00 & 8.20 & 63.89 & 55.25 & 38.00 & 11.20 & 39.90 & 34.42 & 34.46 & \textbf{15.04} \\
& $\mathcal{L}_\text{sup}+\mathcal{L}_\text{unsup}+\mathcal{L}_\text{freq}$ & \textbf{82.24} & \textbf{73.21} & \textbf{31.74} & 9.44 & \textbf{79.83} & \textbf{72.36} & 24.59 & 6.00 & \textbf{68.03} & \textbf{59.55} & \textbf{30.16} & \textbf{8.76} & \textbf{44.84} & \textbf{38.46} & 39.52 & 18.08 \\
\bottomrule
\end{tabular}
}
\end{table}

\begin{table}[!tb]
\centering
\caption{Ablation study on the loss function components using the unseen domains. The best results are \textbf{bolded}.}
\label{tab:ablation_unseen}
\resizebox{0.9\textwidth}{!}{%
\begin{tabular}{l l cccc cccc}
\toprule
\multirow{2}{*}{\textbf{Setting}} & \multirow{2}{*}{\textbf{Design}} & \multicolumn{4}{c}{CVC-300~(Unseen)} & \multicolumn{4}{c}{\texttt{BKAI}~(Unseen)} \\
\cmidrule(lr){3-6} \cmidrule(lr){7-10}
& & Dice$\uparrow$ & Jaccard$\uparrow$ & HD95$\downarrow$ & ASD$\downarrow$ & Dice$\uparrow$ & Jaccard$\uparrow$ & HD95$\downarrow$ & ASD$\downarrow$ \\
\midrule
\multirow{3}{*}{5\% labeled} 
& $\mathcal{L}_\text{sup} + \mathcal{L}_\text{unsup}$ & 76.84 & 69.89 & 21.00 & 9.24 & 56.02 & 46.52 & 43.62 & 17.91 \\
& $\mathcal{L}_\text{sup} + \mathcal{L}_\text{freq}$ & 77.78 & 67.78 & 38.67 & 15.40 & 62.20 & 52.82 & 40.42 & 16.72 \\
& $\mathcal{L}_\text{sup} + \mathcal{L}_\text{unsup} + \mathcal{L}_\text{freq}$ & \textbf{82.46} & \textbf{75.42} & \textbf{21.00} & \textbf{7.09} & \textbf{65.84} & \textbf{56.64} & \textbf{34.69} & \textbf{13.37} \\
\midrule
\multirow{3}{*}{10\% labeled} 
& $\mathcal{L}_\text{sup} + \mathcal{L}_\text{unsup}$ & 84.80 & 77.84 & 16.52 & 6.68 & 61.00 & 52.52 & 33.17 & 13.43 \\
& $\mathcal{L}_\text{sup} + \mathcal{L}_\text{freq}$ & 85.44 & 77.49 & 14.90 & 4.72 & 67.89 & 59.16 & 28.13 & \textbf{10.24} \\
& $\mathcal{L}_\text{sup} + \mathcal{L}_\text{unsup} + \mathcal{L}_\text{freq}$ & \textbf{89.55} & \textbf{81.87} & \textbf{12.30} & \textbf{4.54} & \textbf{68.68} & \textbf{59.91} & \textbf{28.11} & 10.56 \\
\bottomrule
\end{tabular}
}
\end{table}

The complete FPGM model~($\mathcal{L}_\text{sup}+\mathcal{L}_\text{unsup}+\mathcal{L}_\text{freq}$) consistently achieves optimal performance in most metrics, confirming that frequency-guided perturbation and conventional augmentations provide complementary benefits. This synergy is crucial to achieving state-of-the-art performance, with consistent trends across both supervision levels validating the robustness and principled design of our FPGM framework.

\subsubsection{Domain generalization loss component ablation study}

To identify the key contributor to the superior generalization, we performed the same loss component analysis on unseen domains~(Table~\ref{tab:ablation_unseen}). The results provide profound insights: our frequency-guided perturbation~($\mathcal{L}_\text{freq}$) serves as the primary driver of domain generalization capabilities. In both 5\% and 10\% labeled settings, models trained with our proposed perturbation~($\mathcal{L}_\text{sup}+\mathcal{L}_\text{freq}$) consistently and substantially outperform those trained with conventional augmentations~($\mathcal{L}_\text{sup}+\mathcal{L}_\text{unsup}$). This performance gap is particularly striking on the challenging \texttt{BKAI} dataset, where our frequency-guided approach yields remarkable improvements of up to 6.89\% in Dice score.

These findings suggest that by enforcing invariance to a stable, learned structural prior, the model successfully learns intrinsic, generalizable polyp characteristics rather than overfitting to superficial, domain-specific cues. The complete model, which combines both components, achieves optimal generalization performance, indicating that while $\mathcal{L}_\text{freq}$ establishes a robust generalizable foundation, $\mathcal{L}_\text{unsup}$ provides additional photometric robustness benefits.

\subsection{Ablation study on the learned frequency prior}

To validate the core principle of learning frequency priors, we conducted ablation studies comparing our data-driven learned prior against two alternative heuristic, non-learned priors. These alternatives are defined as follows:
\begin{enumerate}[itemsep=2pt,topsep=4pt]
    \item[(i)] Generic Prior: This strategy employs a generic template derived from natural image statistics, specifically a $\frac{1}{f}$ power law distribution~(a.k.a., the pink noise profile). The amplitude at frequency radius $r$ is proportional to $\frac{1}{r+1}$, where $r$ represents Euclidean distance from the zero-frequency component. This approach captures the general tendency for natural images to concentrate energy in lower frequencies.
    \item[(ii)] Heuristic Prior~(Low-pass): This method employs a simple and intuitive heuristic based on an ideal low-pass filter. Specifically, for a given cutoff radius $r_c$, all frequency components within this radius of the direct-current component are preserved, while all components outside this radius are completely suppressed. In our experiments, we set $r_c$=16. This approach tests the hypothesis that simply preserving low-frequency information is sufficient.
\end{enumerate}
The results are unequivocal. As shown in Table~\ref{tab:ablation_frequency_prior}, across both supervision levels and all datasets, our learned prior consistently and substantially outperforms both alternative strategies in region-based metrics~(Dice, Jaccard). This clearly demonstrates that the success of FPGM stems not merely from frequency-space perturbation itself, but critically depends on the quality and relevance of the learned prior.

\begin{table}[!t]
\centering
\caption{Ablation study on the necessity of the learning frequency prior. The best results are \textbf{bolded}.}
\label{tab:ablation_frequency_prior}
\resizebox{\textwidth}{!}{%
\begin{tabular}{l l cccc cccc cccc cccc}
\toprule

\multirow{2}{*}{\textbf{Setting}} & \multirow{2}{*}{\textbf{Design}} & \multicolumn{4}{c}{\texttt{Kvasir}} & \multicolumn{4}{c}{\texttt{CVC-ClinicDB}} & \multicolumn{4}{c}{\texttt{CVC-ColonDB}} & \multicolumn{4}{c}{\texttt{ETIS}} \\
\cmidrule(lr){3-6} \cmidrule(lr){7-10} \cmidrule(lr){11-14} \cmidrule(lr){15-18}

& & Dice$\uparrow$ & Jaccard$\uparrow$ & HD95$\downarrow$ & ASD$\downarrow$ & Dice$\uparrow$ & Jaccard$\uparrow$ & HD95$\downarrow$ & ASD$\downarrow$ & Dice$\uparrow$ & Jaccard$\uparrow$ & HD95$\downarrow$ & ASD$\downarrow$ & Dice$\uparrow$ & Jaccard$\uparrow$ & HD95$\downarrow$ & ASD$\downarrow$ \\
\midrule

\multirow{3}{*}{5\% labeled} 
& Generic prior & 73.87 & 63.58 & 50.87 & 17.40 & 70.48 & 61.47 & 46.18 & 15.17 & 56.22 & 47.18 & 44.61 & 15.51 & 40.57 & 33.33 & 69.66 & 35.33 \\
& Low-pass & 74.43 & 63.82 & 38.86 & 13.10 & 71.11 & 62.65 & 34.59 & 8.61 & 50.40 & 42.71 & \textbf{33.79} & \textbf{11.36} & 44.43 & 38.18 & \textbf{33.41} & \textbf{16.09} \\
& \textbf{Learned prior} & \textbf{80.66} & \textbf{71.34} & \textbf{36.80} & \textbf{12.27} & \textbf{78.01} & \textbf{70.47} & \textbf{27.41} & \textbf{7.46} & \textbf{63.13} & \textbf{54.63} & 40.79 & 13.47 & \textbf{51.07} & \textbf{43.69} & 39.64 & 19.02 \\
\midrule

\multirow{3}{*}{10\% labeled} 
& Generic prior & 80.07 & 70.64 & 36.37 & 10.88 & 76.60 & 69.07 & 27.70 & 7.71 & 62.46 & 54.38 & 36.13 & 12.28 & 40.94 & 35.29 & 35.43 & 16.98 \\
& Low-pass & 81.43 & 72.36 & \textbf{30.31} & \textbf{9.00} & 78.82 & 71.20 & 24.77 & \textbf{5.16} & 61.63 & 53.77 & \textbf{29.07} & 9.37 & 40.32 & 35.69 & \textbf{25.00} & \textbf{11.04} \\
& \textbf{Learned prior} & \textbf{82.24} & \textbf{73.21} & 31.74 & 9.44 & \textbf{79.83} & \textbf{72.36} & \textbf{24.59} & 6.00 & \textbf{68.03} & \textbf{59.55} & 30.16 & \textbf{8.76} & \textbf{44.84} & \textbf{38.46} & 39.52 & 18.08 \\
\bottomrule
\end{tabular}
}
\end{table}

Although generic generic and heuristic low-pass approaches offer modest performance gains over supervised-only baselines, they prove significantly less effective than our learned, task-specific prior. This highlights limitations of using generic or overly simplistic frequency characteristic assumptions. Our approach of learning stable, domain-invariant priors directly from labeled polyp edge information proves far superior, powerfully confirming that data-driven, task-specific priors are essential for optimal performance.

\subsection{Hyperparameter sensitivity analysis}

\begin{figure}[!tb]
\centering
\begin{minipage}[c]{0.49\textwidth}
\includegraphics[width=\textwidth]{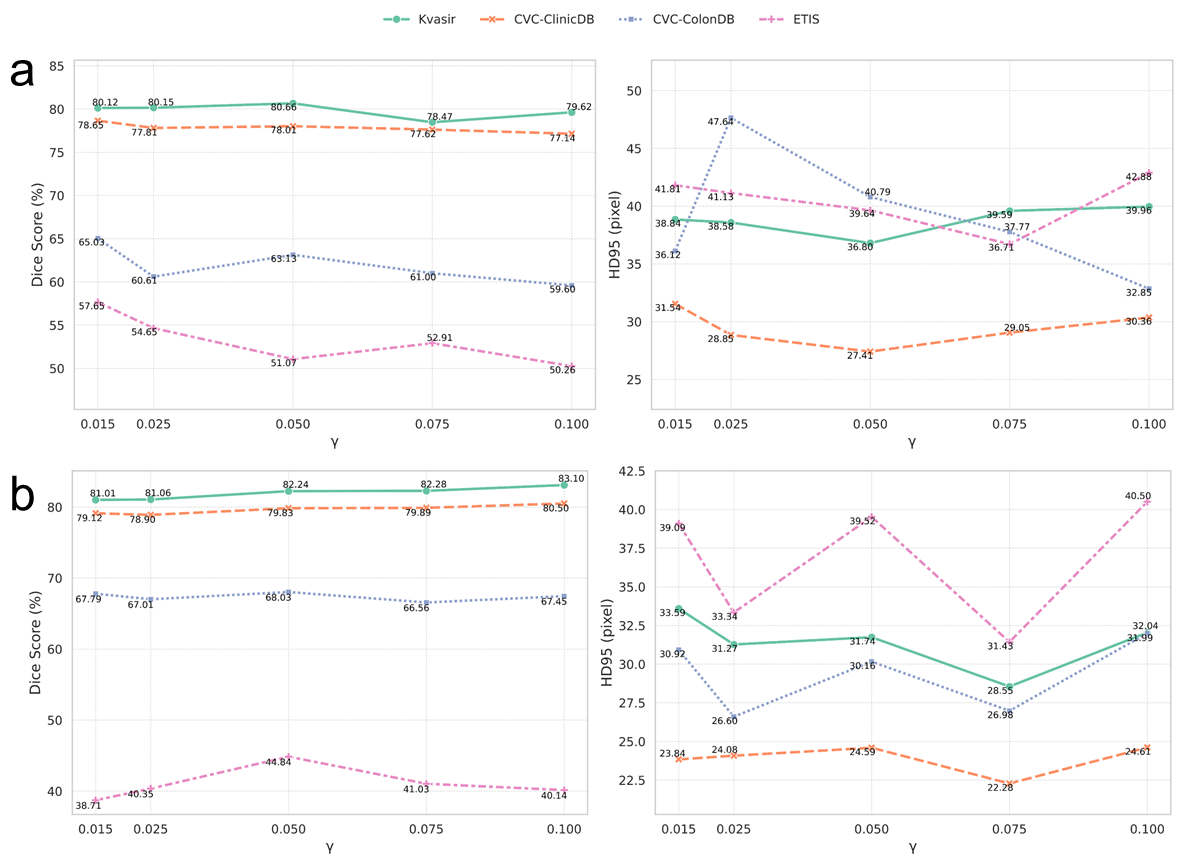}
\caption{Effect of guidance strength $\gamma$ on FPGM segmentation performance. Two core segmentation metrics, Dice Score and HD95, are shown on all four datasets under \textbf{(a)} 5\% and \textbf{(b)} 10\% labeled data settings.}
\label{fig:hyperparam}
\end{minipage}
\hfill
\begin{minipage}[c]{0.49\textwidth}
\includegraphics[width=\textwidth]{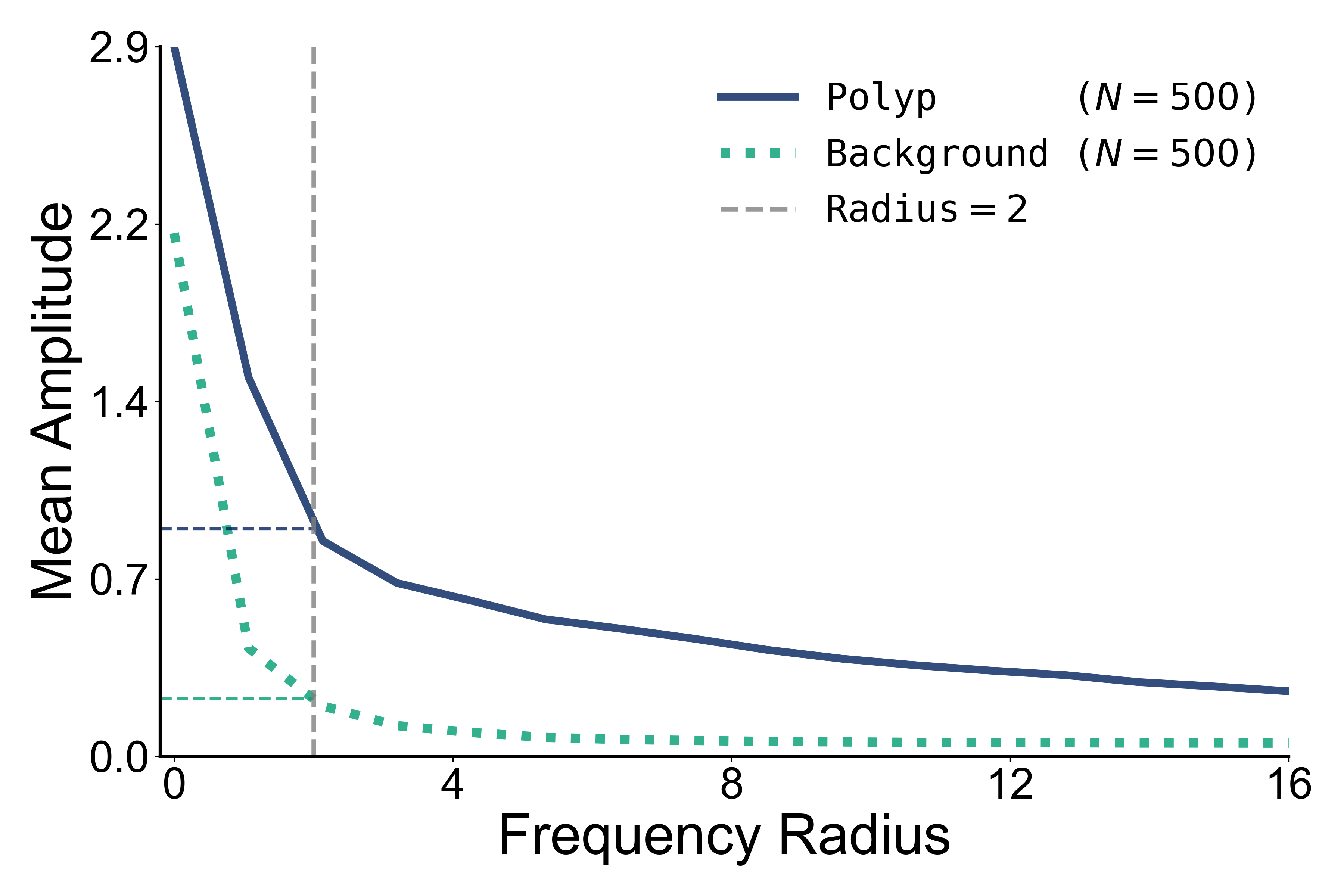}
\caption{Specificity analysis of the learned frequency prior. The average frequency profile of polyp edges is compared against that of an equal number of background pixels, sampled from $N=500$ images. The learned frequency prior is highly specific to the polyp region, while the background profile quickly decays to near-zero amplitude as frequency radius increases.}
\label{fig:prior_background}
\end{minipage}
\end{figure}

We evaluated the robustness of our framework to the key hyperparameter $\gamma$ in Eqn~\eqref{eqn:guided_interpolation}, which controls the strength of spectral shape alignment. The results in Fig.~\ref{fig:hyperparam} illustrate the performance of the model in four datasets in terms of Dice score and HD95 under the settings of 5\% and 10\% labeled data.

The analysis reveals robust performance across a range of $\gamma$ values. In the 5\% setting~(Fig.~\ref{fig:hyperparam}a), Dice scores remain high and stable, and the peak performance is achieved around $\gamma = 0.05$ in most datasets. This indicates that even with extremely limited supervision, our method is not overly sensitive to this parameter.

More nuanced behavior emerges in the 10\% setting~(Fig.~\ref{fig:hyperparam}b). While datasets like \texttt{CVC-ColonDB} and \texttt{ETIS} still favor moderate $\gamma = 0.05$, larger datasets such as \texttt{Kvasir} and \texttt{CVC-ClinicDB} benefit from a slightly stronger perturbation~($\gamma = 0.10$). This suggests that as model proficiency increases with more labeled data, it can tolerate and leverage stronger structural regularization for enhanced performance.

Our selection of $\gamma = 0.05$ as the default value is well justified as it consistently yields optimal or near-optimal performance across all datasets in both scenarios. This validates its use in our main experiments and highlights a desirable property of our framework: achieving state-of-the-art results without requiring extensive, dataset-specific hyperparameter tuning.

\subsection{Specificity analysis of the learned frequency prior}

In order to verify whether the learned frequency prior represents a distinctive signature of polyp pathology rather than generic edge artifacts, we performed a specificity analysis.

We analyzed 500 randomly sampled images from the unified mixed-source training set through controlled comparison. For each image, we isolated polyp edges using ground-truth masks and quantified pixel count $n_\text{polyp}^\text{pixel}$. To ensure a fair comparison, we sampled an identical number of pixels~($n_\text{background}^\text{pixel} = n_\text{polyp}^\text{pixel}$) from the corresponding background regions, defined as areas outside dilated polyp masks to prevent boundary ambiguity. We then computed and aggregated average radial frequency profiles for both polyp edge textures and matched-sample background textures.

Fig.~\ref{fig:prior_background} presents the resulting profiles, revealing a distinct divergence. The polyp edge prior consistently exhibits significantly higher amplitude across the frequency spectrum, indicating a rich and structured textural composition. This is especially obvious at the frequency radius of 2, where both the mixed-source curve and the per-dataset curves~(Fig.~\ref{fig:frequency_signature_combined}b) show non-trivial amplitude at 0.8 to 1.2. In contrast, the background prior remains comparatively flat and near zero at frequency radius over 2, displaying noise-like characteristics.

This finding confirms that our learned prior captures discriminative information that separates polyp edges from arbitrary background textures. This verified specificity provides solid justification for the role of the learned prior within our FPGM framework, explaining its robust generalization capabilities.

\section{Conclusion}
\label{sec:conclusion}

In this paper, we introduce FPGM, a novel data augmentation framework designed to address the critical challenge of semi-supervised polyp segmentation. Moving beyond conventional approaches that rely on generic, random augmentations, we propose a targeted knowledge-transfer paradigm guided by a data-driven frequency prior. This prior, learned from labeled polyp edge regions, encapsulates a stable, discriminative, and domain-invariant structural signature that enables principled, structure-preserving perturbations on unlabeled images.

Our experimental evaluation on six public colonoscopy datasets demonstrates the effectiveness of FPGM in multiple dimensions. The method achieves state-of-the-art performance in standard semi-supervised settings while exhibiting substantially enhanced generalization capabilities on completely unseen datasets compared to ten recent competing approaches. Through comprehensive ablation studies, we confirmed that these performance gains are primarily attributable to our frequency-guided regularization strategy and validated the unique, discriminative nature of the learned structural prior.

The key innovation of FPGM lies in its spectral shape alignment mechanism, which aligns the frequency characteristics of unlabeled images with a canonical polyp representation learned from labeled data. This approach compels the model to focus on truly generalizable features rather than superficial, domain-specific cues. Our findings suggest that frequency-domain knowledge transfer represents a promising direction for medical image analysis, particularly in scenarios where labeled data is scarce and domain generalization is crucial.

While our results are promising, several limitations warrant acknowledgment. The proposed frequency prior is specifically designed and validated for colonoscopy polyp segmentation. Although the underlying principle of learning task-specific structural priors from data is conceptually generalizable, direct applicability to other medical imaging tasks and modalities, such as tumor segmentation in MRI or lesion detection in CT scans, requires further investigation.

Future research directions include exploring whether similar stable and discriminative frequency signatures exist for other anatomical structures or pathologies across different imaging modalities. Additionally, investigating more sophisticated prior modeling approaches, such as learning mixture models to account for different lesion subtypes or exploring alternative spectral transforms beyond the Fourier domain, presents exciting opportunities for advancing structure-aware augmentation strategies in medical imaging.

\section*{Declaration on the Use of AI in Writing}
The authors acknowledge the use of a large language model (Google's large language model) to assist in the language polishing and refinement of this manuscript. The authors meticulously reviewed, edited, and take full responsibility for all content, including the scientific accuracy of the statements and the correctness of all citations. The core ideas, methodologies, and results were generated solely by the authors.

\section*{Acknowledgements}
This work was supported by the National Key Research and Development Program (2022YFC3600902).

\clearpage
\newpage
\bibliographystyle{unsrt}
\bibliography{refs}

\end{document}